\def\preprintversion{1}
\documentclass[10pt,twocolumn,letterpaper]{article}
\ifdefined\preprintversion
  \usepackage[pagenumbers]{cvpr}
\else
  \usepackage[review]{cvpr}
\fi
\usepackage[T1]{fontenc}
\usepackage{amsthm}
\usepackage{microtype}
\graphicspath{{figures/}{../figures/}}
\newtheorem{proposition}{Proposition}
\newcommand{\R}{\mathbb{R}}
\newcommand{\E}{\mathbb{E}}
\newcommand{\ind}{\mathbf{1}}
\newcommand{\relu}{\operatorname{ReLU}}
\newcommand{\canon}{\operatorname{Can}}
\newcommand{\acc}{\operatorname{Acc}}
\newcommand{\dis}{\operatorname{Dis}}
\newcommand{\both}{\operatorname{Both}}

\definecolor{cvprblue}{rgb}{0.21,0.49,0.74}
\usepackage[pagebackref,breaklinks,colorlinks,allcolors=cvprblue]{hyperref}
\usepackage[capitalize,noabbrev]{cleveref}
\hypersetup{pdftitle={Function-Preserving Audits of Visual Weight Readers}}
\def\confName{CVPR}
\def\confYear{2027}
\title{Function-Preserving Audits of Visual Weight Readers}
\ifdefined\preprintversion
\newcommand{\preprintAuthorName}{Sungguk Cha, Mintae Kim, Youngsub Han, Byoung-Ki Jeon, Sangyeob Lee}
\newcommand{\preprintAuthorAffiliation}{LG Uplus, Seoul, South Korea}
\newcommand{\preprintAuthorEmail}{\{sungguk,iammt,yshan042,bkjeon,sangyeob\}@lguplus.co.kr}

  \hypersetup{pdfauthor={\preprintAuthorName}}
  \author{\parbox{\dimexpr\textwidth-2\tabcolsep\relax}{\centering\preprintAuthorName}\\
    \preprintAuthorAffiliation\\
    {\tt\small\preprintAuthorEmail}}
\else
  \hypersetup{pdfauthor={Anonymous Authors}}
  \author{Anonymous Authors}
\fi

\newcommand{\ShapeIoU}{94.6}
\newcommand{\DigitPSNR}{20.7}
\newcommand{\MaxFullError}{1.8\!\times\!10^{-15}}
\newcommand{\MaxRoundedError}{2.0\!\times\!10^{-5}}

\newcommand{\NumLMQueries}{1616}

\begin{document}
\maketitle
\begin{abstract}
A neural field can encode the same image with different weights. A reader
that recognizes one such representation may therefore rely on parameter
coordinates rather than the visual function. We study this distinction
through function-preserving transformations, measuring recognition,
prediction disagreement, and numerical fidelity together. Across shape,
digit, and natural-image fields, high accuracy on the original weights
coexists with substantial sensitivity to equivalent representations.
On 10,000 official CIFAR10 test origins, one-period hidden-bias shifts
change 7.3\% of a released HiddenProbe configuration's predictions while
barely changing accuracy; three-period shifts change 22.0\% and reduce
accuracy from 57.8\% to 54.4\%. Output-only ProbeGen remains unchanged.
These effects are smaller than in a separate adapted-reader study, where
matched phase and sign controls remove the corresponding sensitivities.
Controlled readout studies show why invariance is insufficient: complete
canonical hidden features reach 38.7\% on MNIST, whereas explicitly
composing those activations, outgoing weights, and output bias into field
outputs reaches 81.5\%. The information is present, but the learned
readout does not recover it as effectively.
A separate 720-response study of language models reading numerical
weights shows why stochastic audits require repeats on both inputs:
observed accuracy losses need not yield rejection of response-law equality.
Together, these results distinguish invariance, useful visual information,
and the evidence required to evaluate stochastic readers.
\end{abstract}

\section{Introduction}\label{sec:intro}
Neural fields encode images and geometry as functions of spatial
coordinates \cite{park2019deepsdf,mescheder2019occupancy,chen2021liif}.
Their fitted parameters support visual recognition
\cite{dupont2022functa,zhou2023nfn,adapt2026}, but introduce ambiguities:
neurons can be reordered, ReLU neurons rescaled, and sine biases shifted
by whole periods without changing the signal.

Does a reader recognize the visual function or its parameter coordinates?
Original-weight accuracy cannot distinguish them: fitting regularities
may predict labels while remaining irrelevant to the function, a form of
shortcut learning \cite{geirhos2020shortcut}. Conversely, a constant
classifier can be invariant while recognizing little. Evaluation must
measure both recognition and changes under equivalent representations.

\begin{figure}[t]
\centering\includegraphics[width=\linewidth]{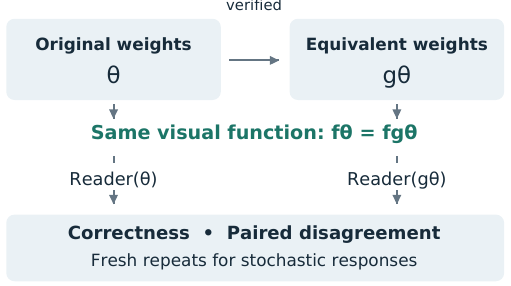}
\caption{\textbf{One visual function, multiple weight inputs.}
A verified transformation changes the parameters while preserving the
field. Paired predictions reveal coordinate dependence; correctness
measures retained visual information. Stochastic readers additionally
require fresh responses to each unchanged input.}\label{fig:overview}
\end{figure}

Our \emph{function-preserving orbit audits} pair original and transformed
predictions, verify the numerical field delivered to the reader, and
retain image identity throughout analysis (\cref{fig:overview}). An orbit
contains the parameterizations reached by the specified field-preserving
transformations.

The audit exposes failures that a single accuracy score conceals.
On adapted CIFAR10 fields, a one-period shift of hidden biases changes 61.6\%
of HiddenProbe predictions even though the checked field discrepancy is
at most $1.12\times10^{-5}$. A matched phase-wrapping control removes
this failure but remains sensitive to sign changes. Features constructed
from sign-invariant products remove both. The sequence of interventions
identifies the parameter information responsible for the observed failure.
An audit of released readers on 10,000 official test origins tests this
behavior beyond the adaptations. It finds smaller accuracy effects but
persistent hidden-reader disagreement, alongside a stable output reader.

Removing that dependence does not determine what information a readout
can use. We therefore compare features taken from field outputs, hidden
activations, and weights. On MNIST, training and evaluating in a common
canonical representation substantially improves the hidden-feature
reader, yet leaves a large gap to reading field outputs. A complete-input control
shows that this gap persists even when the output is exactly recoverable
from the supplied features. Additional
measurements distinguish variability along an exact orbit from variability
between separately fitted fields. This distinction prevents changes in
the represented function from being misattributed to symmetry.

Language models reading numerical weights \cite{takehi2026sparks,kabra2026oracles}
add stochasticity: identical inputs can produce different answers.
Subtracting original-input repeat disagreement is insufficient to test
response-law equality. Repeats on both inputs make an origin-matched
discrepancy estimable; exact conditional tests distinguish accuracy loss
from evidence about the full response law.

The contributions are paired numerical and prediction audits across
three visual populations; targeted phase/sign controls and tests of
canonical readout learning; and a replication design with demonstrated
finite-sample limits for stochastic readers. The transformations and
invariant constructions are established methods. Our contribution is
the controlled comparison and its supported conclusions.

\section{Related Work}\label{sec:related}
\paragraph{Learning visual representations from weights.}
Neural fields represent continuous visual signals
\cite{sitzmann2020siren,tancik2020fourier,dupont2022functa}; their parameters
support prediction through weight statistics \cite{unterthiner2020predicting},
hyper-representations \cite{schurholt2021hyper}, and structured architectures
\cite{survey2026weight}. DWS networks \cite{navon2023dws}, neural functional
networks and transformers \cite{zhou2023nfn,zhou2023nft}, and graph
metanetworks \cite{kofinas2024graphs} encode permutation structure.
Monomial and polynomial functional networks
\cite{tran2024monomial,vo2024polynomial} and ScaleGMN
\cite{kalogeropoulos2024scalegmn} address additional activation-dependent
symmetries. Dayan et al.\ characterize expressivity for both parameter-space
and function-space targets \cite{dayan2026expressivity}. Our audits measure
whether fitted readers retain visual information while respecting these
symmetries.

\paragraph{Fitting and transferring neural representations.}
Fit-a-NeF establishes the importance of initialization, fitting quality,
and capacity for downstream prediction \cite{papa2023fitanef}.
Neural-field adaptation learns structured representations from a shared
base \cite{adapt2026}; HMoE couples an INR classifier with meta-learning
that shapes the field weights \cite{janik2026hmoe}. These methods change
how the representation is learned. Transferable Graph Metanetworks study
width transfer \cite{ma2026transferable}, while the Universal Weight
Encoder uses interface--module graphs to process different architectures
under their supported permutation constraints \cite{kozyra2026interfaces}.
Our audits fix the architecture and fitted visual function during each
symmetry comparison, treating independently refitting a target as a
separate intervention.

\paragraph{Probing a represented network.}
ProbeX applies learned probes directly to selected weight matrices and
routes models to experts \cite{probex2025}. ProbeGen probes final network
outputs \cite{kahana2025probegen}; HiddenProbe also uses intermediate
responses with set-based attention
\cite{sarangi2026hiddenprobe,lee2019settransformer}.
Finite-probe identifiability and universality hold under specified
assumptions on the network family \cite{sarangi2026hiddenprobe}; they do
not establish the invariance or accuracy of every learned readout.
Outputs depend only on the represented function, whereas hidden responses
can change under its symmetries. Our matched phase/sign controls and
feature-location comparisons isolate this distinction.

\paragraph{Task-dependent symmetry and canonicalization.}
The target determines which symmetries a reader should respect: parameter
sensitivity, for example, can vary between functionally identical networks
\cite{kiwitt2025retain}. Here the visual label is unchanged by the audited
actions. Path normalization \cite{neyshabur2015path}, sharpness analysis
\cite{dinh2017sharp}, permutation alignment
\cite{entezari2022permutation,ainsworth2023rebasin,navon2024align}, and
weight augmentation \cite{shamsian2024augment} establish the importance of
parameter coordinates. Canonicalization selects orbit representatives
\cite{kaba2023canonical}, supports weight generation
\cite{deepweightflow2026}, and has known continuity limitations
\cite{dym2024frames}. We evaluate recognition after normalization and
sorting. For sine fields, SIREN-GAP supplies a phasor-graded invariant
reader \cite{sirengap2026software}; our Gram features apply established
sign-invariance principles \cite{lim2023signnet}. We use these principles
to diagnose coordinate dependence and measure the recognition recovered
by a targeted repair.

\paragraph{When does averaging over a symmetry help?}
Test-time averaging can impose symmetry under suitable assumptions.
For non-compact groups, a normalized Haar average is unavailable;
Zhou \emph{et al.} \cite{zhou2026posthoc} study alternative projections
whose benefits depend on the construction and setting. Adly
\cite{adly2026exchange} finds exact agreement between an augmented
predictor with orbit averaging and an equivariant model in a controlled
finite-group experiment. Both hidden-feature and output-based readers
produce predictions that can be averaged. Our finite, sampled actions do
not implement an exact group projection, and their effect on recognition
can have either sign. We measure this effect alongside coordinate
sensitivity without treating the latter as a predictor of accuracy gain.

\paragraph{Language-model and distributional evaluation.}
Weight-space programming and Weight Oracles study language models that
read or manipulate network weights \cite{takehi2026sparks,kabra2026oracles}.
We audit frozen readers using verified metamorphic relations
\cite{segura2016metamorphic,ribeiro2020checklist}. Numeracy and prompt/answer
biases motivate matched interface controls
\cite{wallace2019numeracy,sclar2024format,lu2022order,zheng2024selectors,
zhao2021calibrate}, with generated answers separated from token choices
\cite{wang2024text}. Our distributional analysis builds on MMD
\cite{gretton2012kernel}, model equality testing \cite{gao2025equality},
conditional two-sample tests \cite{chatterjee2024conditional}, and group
randomization \cite{chiu2025symmetry}. RUT uses reference-model token
scores to detect API substitutions with one target response per prompt
\cite{zhu2026rut}. Such tests can detect particular alternatives with a
single response. Our replication analysis concerns uniform guarantees
over unrestricted origin-specific laws \cite{aliakbarpour2020multiple}.
Two-sided repeats estimate the full categorical discrepancy without
reference logits.

\section{Function-Preserving Orbit Audits}\label{sec:method}
\subsection{Hold the visual function fixed}
Let $f_\theta$ be a neural field and $H(\theta)$ a reader that predicts the
label of the visual target encoded by its parameters. A transformation
$g$ is admissible when $f_{g\theta}=f_\theta$ in exact arithmetic.
The visual label is unchanged by these actions; parameter-dependent
targets can require different invariances \cite{kiwitt2025retain}.
The independent evaluation unit is the \emph{visual origin}: a shape or
source image, carrying all of its fitted and transformed representations.
Origins, rather than transformed copies, are assigned to data splits.

For a scalar ReLU field with one hidden layer,
\begin{equation}
 f_\theta(x)=\sum_{j=1}^{h}v_j\relu(w_j^\top x+b_j)+c,
 \qquad x\in[-1,1]^2,\label{eq:field}
\end{equation}
positive homogeneity permits the action
\begin{equation}
 (w'_j,b'_j,v'_j)
 =(s_jw_{\pi(j)},s_jb_{\pi(j)},v_{\pi(j)}/s_j),\quad s_j>0.
 \label{eq:action}
\end{equation}
We evaluate permutations, scales, and their composition. The primary
scales are $s_j=2^{k_j}$ with $k_j$ uniform on $\{-4,\ldots,4\}$.
This choice gives exact binary rescaling of the stored weights within
their numerical range. Scale-only and joint conditions use separate draws.

For a SIREN layer
$h_\ell=\sin(\omega(W_\ell h_{\ell-1}+b_\ell))$,
periodicity permits $b_{\ell j}\mapsto b_{\ell j}+2\pi k/\omega$.
Negating an incoming affine row and the corresponding outgoing column
also preserves the field, as does a compensated neuron permutation
\cite{shamsian2024augment}. We use independently signed one- and
three-period shifts, denoted Phase 1 and Phase 3. ReLU rescaling is not
a sine-network symmetry. Refitting a target is also a different
intervention: it generally changes the visual function.

\subsection{Measure correctness and paired changes}
Write $Y_0$ and $Y_g$ for the original and transformed predictions, and
$y$ for the target label. The audit reports
\begin{align}
 \acc_g &= \E\ind\{Y_g=y\},\nonumber\\
 \dis_g &= \E\ind\{Y_0\ne Y_g\},\nonumber\\
 \both_g &= \E\ind\{Y_0=Y_g=y\}.\label{eq:metrics}
\end{align}
For behavioral scoring, invalid answers are errors and any pair containing
one counts as disagreement. If $r_g$ is the correct-to-incorrect rate,
$u_g$ the incorrect-to-correct rate, and $q_g$ the rate of disagreement
between two incorrect answers, then
\begin{equation}
 \acc_g-\acc_0=u_g-r_g,\qquad
 \dis_g=r_g+u_g+q_g.\label{eq:decompose}
\end{equation}
Accuracy changes can therefore conceal many prediction changes. Reporting
both quantities also distinguishes useful invariance from constant output.

For deterministic readers with probability vector $P_{ig}$, we additionally
measure variation across a fixed set of transformations. Define
\begin{align}
 V_g &= \E_i\E_g\|P_{ig}-\bar P_i\|_2^2,\nonumber\\
 V_o &= \E_i\|\bar P_i-\bar P\|_2^2,\qquad
 G_o=\frac{V_g}{V_g+V_o}.\label{eq:ges}
\end{align}
Here $\bar P_i=\E_gP_{ig}$ and $\bar P=\E_i\bar P_i$.
This label-free \emph{orbit-variance share} describes the chosen
transformation design (zero for a constant probability vector); it does
not predict whether averaging transformed
predictions will improve accuracy. For multiple fitting sources, a
separate ratio $G_f=V_g/(V_g+V_f)$ compares orbit variability with
between-fit variability, as defined in the supplement. The two ratios
have different denominators and are reported separately.

\subsection{Construct targeted invariant controls}
For the ReLU field, let $a_j=(w_j,b_j)$ and $r_j=\|a_j\|_2$.
Normalize nonzero affine rows by
\begin{equation}
 (\bar a_j,\bar v_j)=(a_j/r_j,r_jv_j),\label{eq:normalize}
\end{equation}
and sort the complete normalized neuron tuples lexicographically.
Map zero affine rows to zero tuples and retain $c$. The resulting map
$\canon$ satisfies
\begin{equation}
 \canon(g\theta)=\canon(\theta),\qquad
 f_{\canon(\theta)}=f_\theta.\label{eq:canonical}
\end{equation}
The outgoing weight resolves ties between identical affine rows.
At fixed input dimension, the construction costs $O(h\log h)$.
It selects representatives for \cref{eq:action}, not for every possible
functional equivalence. A reader trained on canonical inputs must also
be evaluated on canonical inputs; applying this map only at test time
introduces a separate distribution shift.

For sine networks, our first matched control, \emph{HiddenPhase}, wraps
each hidden preactivation $z$ to
$\operatorname{atan2}(\sin\omega z,\cos\omega z)/\omega$.
This removes whole-period phase shifts but retains sign dependence.
The second, \emph{HiddenGram}, reads the post-sine profile
$h\in\R^{128}$ across learned probes through
$\Phi(h)=(h_1h,h_2h,\nu(h))$, where $\nu$ contains seven even statistics.
Because $\Phi(-h)=\Phi(h)$, compensated negations leave these features
unchanged. Permutation-invariant aggregation then handles neuron order.
Both controls retain the original HiddenProbe head and its 447,908
parameters. These constructions test a specific mechanism: which hidden
coordinates the readout needs to discard.

\subsection{Verify the numerical intervention}
A symbolic symmetry does not guarantee identical finite-precision inputs.
We execute the transformed fields on fixed audit coordinates, including
learned probe locations for SIRENs. For language-model inputs, we also
execute the actual six-significant-digit decimal weights.
Let $S_p$ denote decimal serialization and parsing. We measure
\begin{equation}
 \epsilon_p=\max_{x\in Q}
 |f_{S_p(\theta)}(x)-f_{S_p(g\theta)}(x)|,\label{eq:rounding}
\end{equation}
and each representation's discrepancy from its full-precision origin.
The ReLU audit additionally gives a uniform bound on $[-1,1]^2$.
For paired reference and displayed parameters, put
$M_j=\max(0,b_j+\|w_j\|_1)$ and
$\delta_j=\|w_j-\widehat w_j\|_1+|b_j-\widehat b_j|$. Then
\begin{equation}
 \|f-\widehat f\|_\infty\leq |c-\widehat c|
 +\sum_j\bigl(|v_j-\widehat v_j|M_j+|\widehat v_j|\delta_j\bigr).
 \label{eq:uniform}
\end{equation}
This follows from ReLU's one-Lipschitz property. Exact rational evaluation
bounds the fields defined by the printed numbers; sign is preserved
where the reference magnitude exceeds the bound. Finite-query SIREN
checks are numerical diagnostics, not uniform certificates.

\subsection{Use two-sided replication for stochastic readers}
\label{sec:repeats}
For generated answers, let $p_i$ and $q_i$ denote the class-response laws
of the original and transformed input at origin $i$. Independent original
repeats provide a baseline, but subtracting it gives
\begin{equation}
 D(p,q)-D(p,p)=\|p\|_2^2-p^\top q,\label{eq:excess}
\end{equation}
where $D(p,q)=1-p^\top q$. This is not a distributional distance: if
$p$ is uniform, it vanishes for every $q$. Moreover, one transformed
response per origin cannot support a uniformly powerful test of equality
for unrestricted origin-specific laws, even with unlimited original
repeats \cite{aliakbarpour2020multiple}.

With $m\geq2$ independent responses per side, we can instead estimate
\begin{equation}
 S_n=\frac{1}{2n}\sum_{i=1}^{n}\|p_i-q_i\|_2^2.\label{eq:symmetric}
\end{equation}
For $X_{i1:m}\sim p_i$ and $Z_{i1:m}\sim q_i$, use
\begin{align}
 T_i^{(m)}={}&\frac{1}{2\binom m2}\sum_{a<b}
 \left(\ind\{X_{ia}=X_{ib}\}+\ind\{Z_{ia}=Z_{ib}\}\right)\nonumber\\
 &-\frac{1}{m^2}\sum_{a,b}\ind\{X_{ia}=Z_{ib}\}.
 \label{eq:two_repeat}
\end{align}
The within-input collision expectations are $\|p_i\|^2$ and
$\|q_i\|^2$; cross-input collisions have expectation $p_i^\top q_i$.
Thus $\widehat S_n=n^{-1}\sum_iT_i^{(m)}$ is unbiased for $S_n$, an
origin-matched categorical MMD discrepancy \cite{gretton2012kernel,gao2025equality}.
For this statistic an invalid answer is a separate categorical symbol.

Under equality and independent, stationary calls, condition labels are
exchangeable within each origin. Enumerating all balanced allocations
and convolving their integer-valued statistics gives an exact upper-tail
test. There are six allocations for two responses per side and twenty
for three. The supplement derives the replication limit and the exact
null. These results identify an estimable target and a valid test; neither
guarantees adequate power in a small experiment.

\begin{table*}[t]
\centering\small
\setlength{\tabcolsep}{5pt}
\caption{\textbf{Recognition and coordinate sensitivity on ReLU fields.} Accuracy and raw/joint disagreement are percentages on 120 shape and 200 digit test origins. Joint transformations combine neuron permutation and positive rescaling. The probe reader queries the represented function; other rows read parameters.}\label{tab:relu_summary}
\begin{tabular*}{\textwidth}{@{\extracolsep{\fill}}lrrrrrr@{}}
\toprule
& \multicolumn{3}{c}{Shapes} & \multicolumn{3}{c}{MNIST} \\
\cmidrule(lr){2-4}\cmidrule(lr){5-7}
Reader & Raw & Joint & Disagree & Raw & Joint & Disagree \\
\midrule
Flat ridge & 32.5 & 20.0 & 75.0 & 13.5 & 7.0 & 89.5 \\
Stats ridge & 96.7 & 24.2 & 75.0 & 33.0 & 11.5 & 89.0 \\
Norm+Stats & 97.5 & 97.5 & 0.0 & 33.5 & 33.5 & 0.0 \\
Canon+Flat & 71.7 & 71.7 & 0.0 & 33.5 & 33.5 & 0.0 \\
Probe ridge & 100.0 & 100.0 & 0.0 & 74.0 & 74.0 & 0.0 \\
Set & 96.7 & 20.0 & 80.8 & 68.5 & 20.0 & 83.5 \\
Set+Aug & 100.0 & 95.0 & 5.0 & 66.0 & 55.5 & 38.0 \\
Norm+Set & 95.0 & 95.0 & 0.0 & 69.0 & 69.0 & 0.0 \\
\bottomrule
\end{tabular*}
\end{table*}

\section{Experimental Design}\label{sec:setup}
\paragraph{Visual populations.}
The ReLU study contains 480 fields of six shape classes and 1,000 MNIST
fields \cite{lecun1998mnist}. Shapes use width 32 and a 300/60/120
training/validation/test split; digits use width 64 and a 600/200/200
split. All fields receive 6,000 fitting updates. Mean reconstruction
quality is \ShapeIoU\% IoU for shapes and \DigitPSNR\,dB for digits.
Every fit is retained. Language-model readers use a fixed, balanced subset
of 60 shape test origins.

The natural-image orbit study uses 1,600 CIFAR10 originals
\cite{krizhevsky2009cifar}: 1,000 for training, 200 for validation, and
400 from the official test partition. Three independently initialized
fitting bundles each supply two sources, totaling 9,600 SIRENs with
architecture $2$--$64$--$64$--$3$ and frequency $\omega=8$.
Each source shares an initialization across images. Reader training uses
one source; the second is reserved for transfer evaluation.

\paragraph{Readers and selection.}
The ReLU comparisons include flat and statistical ridge readouts,
DeepSets readers \cite{zaheer2017sets}, their normalization or augmentation
controls, and an 81-query ridge reader of field outputs. Set-reader
predictions average three initialization seeds. The natural-image study
compares HiddenProbe, ProbeGen, MPGMN, fixed-grid readers, weight
statistics, a reconstructed-image CNN, and phase, sign, augmentation,
and phasor controls. Three reader seeds per fitting bundle give 99
validation-selected states. These states use the same 240,000-image presentation
budget and are fixed before test evaluation. Architectural and
optimization differences remain outside the explicitly matched controls.

A released-configuration follow-up uses the official 60,000-field CIFAR10
corpus ($2$--$32$--$32$--$3$, $\omega=30$), with 45k/5k/10k splits and
three seeds each for HiddenProbe and ProbeGen \cite{sarangi2026hiddenprobe}.
We preserve the pinned released 60-epoch training recipes and clean
validation selection. The trainers log test scores during training;
these do not drive selection. The audit uses FP64 targets/actions and
FP32 readouts, with native clean evaluation reported separately.

\paragraph{Feature-location and fitting diagnostics.}
Separate retrospective studies use the earlier 600-image CIFAR10 pilot
and the MNIST fields. They compare output, hidden, and weight features
with a common two-hidden-layer readout and optimization schedule.
Five seeds evaluate feature location; six evaluate canonical training
on MNIST. A six-seed follow-up adds complete hidden inputs and an exact
output-composition control, replaying all five readouts on CPU. Input
dimensions differ and are documented in the supplement.
A 42-state pilot analysis compares orbit and fitting-source variability.
A separate low-pass intervention tests how target bandwidth changes
agreement between independent SIREN fits.

\paragraph{Stochastic readers.}
The prompt defines the field, foreground convention, six shape classes,
and all numerical weights. A randomized answer menu is shared across
conditions for each origin. Four requested model configurations supply
1,920 exploratory generated answers, including raw and canonical repeats.
A separate prospective study selects the two stronger original-weight
recognizers, Astra and Sol, and collects three fresh raw and three fresh
joint responses per origin: 720 answers. This study fixes the prompts,
order randomization, inclusion rule, and two-test Holm family
\cite{holm1979simple} before collecting responses. Historical answers
are not pooled into it. Deterministic next-token models, redacted weights,
and explicit shape cues provide separate interface controls.

\paragraph{Paired uncertainty.}
Rates use Wilson intervals \cite{wilson1927interval}; paired changes use
class-stratified origin bootstraps \cite{efron1979bootstrap}. All conditions,
reader seeds, and fitting bundles travel with their original image.
The original ReLU and generated-answer analyses use 2,000 resamples;
the prospective studies use 10,000. Intervals condition on trained reader
states. Exploratory feature-location results report seed means; repeated
predictions of one image are not treated as independent images.
The supplement provides complete protocols, response-distribution tests,
and the scope of retained artifacts.

\section{Results}\label{sec:results}
\begin{figure*}[t]
\centering\includegraphics[width=.98\textwidth]{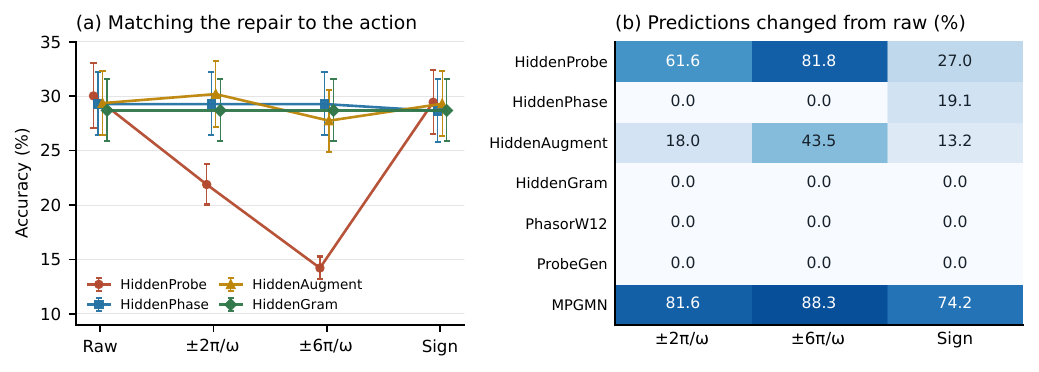}
\caption{\textbf{Adapted natural-image readers: orbit sensitivity and targeted controls.}
All conditions preserve the SIREN function in real arithmetic.
(a) Accuracy on 400 CIFAR10 originals, averaged over three fitting bundles
and three reader seeds; bars are 95\% original-image bootstrap intervals.
Phase wrapping removes phase sensitivity; Gram features also remove sign
sensitivity. (b) Paired prediction disagreement exposes changes concealed
by similar net accuracy. The existing phasor reader and output-only probes
are additional invariant controls.}\label{fig:orbit}
\end{figure*}

\subsection{Accurate recognition can depend on arbitrary coordinates}
The ReLU results establish the central failure in a setting where readers
recognize the original inputs well (\cref{tab:relu_summary}). Shape Stats
ridge falls from 96.7\% to 24.2\% under joint permutation and rescaling:
87 of 120 correct predictions become incorrect and none improve.
The paired loss is 72.5 points, with a 95\% interval of
$[66.7,78.3]$. The permutation-invariant set reader falls from 96.7\%
to 20.0\%. Ignoring neuron order therefore leaves substantial dependence
on neuron scale.

Normalization preserves 97.5\% accuracy for statistics and 95.0\% for
the set reader, whereas equally invariant canonical flat features reach
only 71.7\%. The output-query reader reaches 100.0\% in every condition.
These controls show both that the fitted functions retain the label
information and that an invariance guarantee does not determine how
well a readout extracts it.

MNIST gives the same distinction on a harder recognition task. The set
reader falls from 68.5\% to 20.0\%, changing 83.5\% of predictions.
Augmentation improves joint accuracy to 55.5\% but leaves 38.0\%
disagreement. Its 38 harmful and 17 helpful flips partially cancel in
the 10.5-point net loss. Normalized set features preserve 69.0\%,
while normalized statistics and canonical flat features both reach
33.5\%. The strongest-reconstruction quartiles retain large losses,
so poor fitting alone does not explain these results.

\subsection{Phase and sign controls localize natural-image failures}
\label{sec:orbit_results}
The adapted natural-image study tests whether the same behavior occurs in
modern readers of deeper fields. Across the six source/bundle
combinations, mean reconstruction PSNR ranges from 31.4 to 32.6\,dB.
Function checks give maximum deviations of $1.75\times10^{-14}$ for
ideal FP64 actions and $1.12\times10^{-5}$ after FP32 rounding,
including the learned probe coordinates.

\paragraph{Removing phase dependence.}
HiddenProbe falls from 30.0\% original accuracy to 21.9\% after one-period
hidden-bias shifts (\cref{fig:orbit}). The primary paired change is
$-8.14$ points, with a 95\% interval of $[-10.36,-5.97]$ and
Holm-adjusted exact sign-randomization $p=2.23\times10^{-8}$ under the
specified exchangeability null. Every fitting bundle shows a loss:
9.75, 6.17, and 8.50 points. Three-period shifts reduce accuracy further
to 14.2\%. These shifts leave the visual function unchanged but move
the preactivations supplied to the readout.

Wrapping those preactivations to one period removes the phase failure.
HiddenPhase retains 29.3\% accuracy with no observed prediction changes
under either phase condition. Its clean difference from HiddenProbe is
$-0.75$ points, with interval $[-2.47,1.03]$. The matched control
identifies phase dependence while quantifying its recognition cost;
the interval alone does not establish accuracy equivalence.

\paragraph{Testing the remaining symmetry.}
Phase wrapping still changes 19.1\% of predictions under compensated
neuron negation, despite a net accuracy loss of only 0.64 points.
HiddenGram removes this sign dependence as well as phase dependence,
retaining 28.7\% accuracy and unchanged predictions under every tested
action on both sources. Its clean difference from HiddenProbe is
$-1.33$ points $[-3.08,0.42]$. The construction uses known invariant
products and the same parameter count, providing a direct control for
which hidden information reaches the reader.

Training with phase augmentation gives a different outcome: phase-one
accuracy remains 30.2\%, but predictions change on 18.0\% of comparisons;
phase-three disagreement reaches 43.5\%. The tested training distribution
therefore gives partial robustness without the structural guarantee.
All methods have zero observed permutation disagreement, showing that
permutation handling alone does not resolve phase or sign dependence.

\paragraph{Recognition after invariance.}
The existing phasor reader also has unchanged predictions, at 24.5\%
original accuracy. ProbeGen reaches 29.2\%, Grid64 34.9\%, and the
reconstructed-image CNN 44.1\%. MPGMN changes from 25.2\% to 12.9\%
under Phase 1. These comparisons place the targeted repairs in context:
removing a coordinate dependence is useful, but does not by itself make
hidden-weight reading the strongest way to recognize these images.

\subsection{Released readers retain sensitivity at benchmark scale}
\label{sec:official_cifar}
\begin{table}[t]
\centering\small\setlength{\tabcolsep}{3pt}
\caption{\textbf{Official CIFAR10 corpus audit.} Accuracy (\%) on 10,000 test origins, mean of 3 independently trained readers per method. $D_1$ is Phase-1 prediction disagreement (\%). Targets/actions use FP64; readouts use FP32. Native clean precision and paired intervals are in the supplement.}\label{tab:official_cifar}
\begin{tabular}{lrrrrr}\toprule
Reader & Raw & Phase 1 & Phase 3 & Sign & $D_1$\\\midrule
HiddenProbe & 57.8 & 57.6 & 54.4 & 57.9 & 7.3 \\
ProbeGen & 57.3 & 57.3 & 57.3 & 57.3 & 0.0 \\
\bottomrule\end{tabular}
\end{table}

On the official corpus, native clean accuracy is 57.82\% for HiddenProbe
and 57.28\% for ProbeGen (three-seed standard deviations 0.40 and 0.62
points). The FP64-target audit gives HiddenProbe 57.84\% clean accuracy.
Every transformed field passes the numerical gate, with maximum raw
function error $1.52\times10^{-14}$ across grid and learned-probe checks.

The released configuration attenuates the adapted study's Phase-1
accuracy loss: the change is $-0.23$ points, with descriptive paired
95\% interval $[-0.48,0.02]$. Yet 7.28\% of predictions change
(\cref{tab:official_cifar}). Phase 3 changes 22.03\% and reduces accuracy
by 3.41 points $[-3.88,-2.95]$. Sign flips change 10.52\% despite a
net gain of 0.10 points $[-0.20,0.40]$. Thus small accuracy changes
can conceal coordinate dependence. ProbeGen and permutation controls
have no observed prediction changes. The effect size varies between
configurations; the large adapted-study loss is not universal.

Precision is part of the intervention: upstream BF16 inference can
change phase-shifted field outputs substantially through rounding.
The audit therefore compares clean and transformed targets in FP64
with the same frozen FP32 readout, rather than mixing inference
pipelines. Native scores, operator checks, and per-seed results are
reported in the supplementary official-corpus protocol.

\subsection{Canonicalization repairs sensitivity, not readout learning}
\label{sec:taps}
\begin{table*}[t]
\centering\small
\setlength{\tabcolsep}{5pt}
\caption{\textbf{Feature location changes orbit sensitivity and the effect of averaging.} Means over five reader seeds on 100 independent CIFAR10 audit origins or 200 independent MNIST test origins. Seeds and transformed copies are repeated measurements, not additional independent images. Averaged predictions use the recorded 16-entry transformation design; CIFAR10 entries repeat four fixed actions. Input dimensions differ across feature types, so rows are not capacity-matched. $G_o$ is the orbit-variance share in Eq.~\eqref{eq:ges}; changes are percentage points.}\label{tab:tap_summary}
\begin{tabular*}{\textwidth}{@{\extracolsep{\fill}}llrrrr@{}}
\toprule
Population & Reader features & Raw (\%) & Averaged (\%) & Change & $G_o$ \\
\midrule
CIFAR10 & Field outputs & 24.2 & 24.2 & +0.0 & $<10^{-9}$ \\
CIFAR10 & Hidden activations & 16.8 & 18.4 & +1.6 & 0.644 \\
CIFAR10 & Hidden preactivations & 21.8 & 11.4 & -10.4 & 0.774 \\
CIFAR10 & Weight statistics & 17.8 & 10.6 & -7.2 & 0.916 \\
\midrule
MNIST & Field outputs & 81.1 & 81.1 & +0.0 & $<10^{-9}$ \\
MNIST & Hidden activations & 14.3 & 12.5 & -1.8 & 0.893 \\
MNIST & Hidden preactivations & 11.2 & 11.5 & +0.3 & 0.890 \\
MNIST & Weight statistics & 31.6 & 12.9 & -18.7 & 0.862 \\
\bottomrule
\end{tabular*}
\end{table*}

We next vary where the reader extracts its features
(\cref{tab:tap_summary}). Field-output features have negligible orbit
variance and unchanged accuracy after averaging predictions across
transformations. Hidden and weight features expose coordinates that the
complete field eliminates. On MNIST, averaging reduces the weight-statistic
reader from 31.6\% to 12.9\%; on CIFAR10 it reduces the hidden-preactivation
reader from 21.8\% to 11.4\%. The outcome is not uniformly negative:
CIFAR10 post-sine features gain 1.6 points. Large orbit variability is
a diagnostic of sensitivity, not a guarantee of either benefit or harm
from averaging.

\begin{figure}[t]
\centering\includegraphics[width=.86\linewidth]{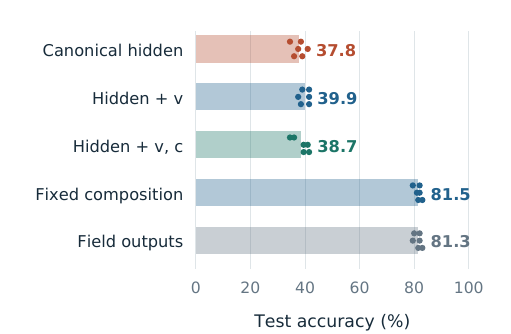}
\caption{\textbf{Complete information still requires an effective readout.}
MNIST follow-up, six seeds on 200 test images; dots are seeds and bars
are means. Hidden features are canonical during training and evaluation.
$v,c$ denote outgoing weights and bias. Fixed composition reconstructs
field outputs before applying the output reader. Input dimensions and
parameter counts differ.}\label{fig:rescue}
\end{figure}

Canonical training addresses coordinate mismatch: in the earlier
six-seed MNIST study it raises hidden-feature accuracy from 15.2\% to
38.5\%. A new six-seed control separates recoverable information from
the ability to use it (\cref{fig:rescue}). Complete canonical profiles,
outgoing weights $v$, and output bias $c$ yield 38.7\%, versus 39.9\%
without $c$. Explicitly computing $\sum_jv_jh_j(x)+c$ from those same
complete features recovers field outputs exactly and reaches 81.5\%,
versus 81.3\% for direct outputs in this run. The information is therefore
present, but the tested learner does not exploit it comparably. Unequal
input dimensions and parameter counts prevent a capacity-isolated
interpretation; paired origin intervals are in the supplement.

The 42-state fitting-source comparison reinforces the distinction.
All 24 output-based states have orbit-to-fit variance ratio
$G_f<10^{-9}$ and unchanged averaged predictions. The 18 hidden/weight
states have $G_f$ between 0.848 and 0.985, yet their averaging effects
have both signs. These ratios describe the measured source and
transformation design; they do not establish a universal threshold for
selecting a reader or enabling test-time averaging.

\subsection{Refitting changes a different part of the problem}
\label{sec:refitting}
Exact orbits preserve the field, whereas independently fitting the same
image can change it. In the 192-image bandwidth intervention, independent
SIREN fits of the original targets agree at 35.00\,dB, while their mean
reconstruction PSNR is 32.09\,dB, both in RGB units. Retaining one
quarter of the radial Nyquist frequency raises agreement to 50.80\,dB
but lowers fidelity to the original images to 20.95\,dB. Smoothing therefore makes
fitting more repeatable while discarding visual detail.

The separate paired reader study, restricted to its 100 audit images,
finds mean accuracy of 27.0\% for the original targets and 24.3\% at the strongest low-pass setting.
Improved agreement between fits does not translate into an observed
recognition gain in this experiment. Since smoothing also changes the
target, this intervention cannot attribute all residual reader error
to high-frequency ambiguity. Complete bandwidth results and the
independent source-transfer pilot appear in the supplement.

\subsection{Stochastic audits require repeats on both inputs}
\label{sec:generated_results}
Three independent responses per raw and joint input on 60 origins
give Astra 88.3\% versus 81.1\% accuracy and Sol 86.7\% versus 79.4\%.
Both descriptive accuracy-loss intervals exclude zero. However,
original-input repeats already disagree on 13.3\% and 16.7\% of pairs,
compared with 15.7\% and 20.9\% cross-input disagreement. Both exact
response-distribution tests give Holm-adjusted $p=0.162$.
Nonrejection does not establish invariance: only 9 and 13 origins
inform the conditional statistic. Replication makes the discrepancy
estimable; the response patterns determine finite-sample evidence.
Full tables, paired accuracy intervals, and the earlier 1,920-response
exploration are in the supplement.

\subsection{Numerical and interface controls support the interpretation}
The maximum full-precision ReLU discrepancy is $\MaxFullError$.
At the primary six-digit precision, the maximum sampled rounded-pair
error is $\MaxRoundedError$. Exact-rational bounds limit the raw/joint
field difference uniformly to $7.45\times10^{-5}$ for shapes and
$4.82\times10^{-4}$ for MNIST. Executing the printed inputs changes at
most one foreground indicator among 4,096 audit coordinates per origin.
These checks quantify the numerical perturbation accompanying the
reader failures.

Next-token controls reinforce the need for a recognition reference:
Qwen2.5-0.5B gives the same letter on all 60 original and redacted inputs.
A Qwen3.5-27B control reads 7/60 weight inputs but 60/60 explicit cues.
Known-law simulations illustrate low finite-sample power; full interface
and replication results are in the supplement.

\section{Discussion and Conclusion}\label{sec:discussion}\label{sec:conclusion}
Function-preserving transformations reveal whether a reader relies on
coordinates that the visual function discards. The natural-image controls
identify a concrete sequence: phase wrapping removes periodic bias
dependence, while invariant products also remove sign dependence.
The ReLU and feature-location studies establish the complementary result:
even a complete, canonical hidden representation can be difficult for
a learned readout to use, although explicit composition recovers outputs. Invariance and useful information must be evaluated together.

Evaluation should verify numerical preservation, report prediction
changes alongside recognition, include an output-query reference, and
match training and test coordinates when canonicalizing. Stochastic
audits additionally need repeats on both inputs and an explicit target:
accuracy, disagreement, or response-law equality.

The scope remains shallow ReLU and two-hidden-layer SIREN fields.
Mechanistic repairs use adapted readers; the official follow-up confirms
sensitivity with smaller accuracy effects in one released configuration
per method. Feature-location comparisons have unequal capacity, and
smoothing changes the target. Generated-answer tests assume independent
stationary calls and have limited power at three repeats per side.

\clearpage
\section*{LLM Usage Statement}
OpenAI Codex assisted research design, literature discovery, implementation,
experimental execution, analysis, manuscript preparation, and plotting-code
development. Figures are programmatic visualizations of measured data.
The language models evaluated as readers and their inference interfaces are
specified in the experimental setup and supplementary methods.
{\small\bibliographystyle{ieeenat_fullname}\bibliography{references}}
\ifdefined\preprintversion
\clearpage\appendix
\renewcommand{\topfraction}{.95}
\renewcommand{\dbltopfraction}{.95}
\renewcommand{\textfraction}{.05}
\renewcommand{\floatpagefraction}{.8}
\renewcommand{\dblfloatpagefraction}{.8}
\setcounter{topnumber}{8}
\setcounter{dbltopnumber}{8}
\setcounter{totalnumber}{12}
\section{Populations, Targets, and Evaluation Units}
\label{app:targets}
The audit follows visual identities through every transformation and refit.
The ReLU evaluation sets contain 120 shape origins and 200 MNIST images;
language-model readers share a balanced subset of 60 shape origins.
The natural-image orbit study uses 400 official CIFAR10 test images.
Repeated representations and reader seeds add paired measurements, not
independent visual examples. All reported analyses retain imperfect fits.

\subsection{Geometric fields}
Let $z$ be a translated, rotated, independently axis-scaled coordinate,
with $a=|z_1|$, $b=|z_2|$, and $r=\|z\|_2$. The six implicit targets are
\begin{align*}
 \text{circle}:&\quad r-0.58,\\
 \text{square}:&\quad\max(a,b)-0.52,\\
 \text{diamond}:&\quad(a+b-0.78)/\sqrt{2},\\
 \text{triangle}:&\quad\max_{n\in N}n^\top z-0.34,\\
 \text{cross}:&\quad\min\{\max(a-0.20,b-0.65),\\
 &\hspace{45pt}\max(a-0.65,b-0.20)\},\\
 \text{ring}:&\quad|r-0.49|-0.17,
\end{align*}
where $N=\{(0,-1),(\sqrt3/2,1/2),(-\sqrt3/2,1/2)\}$.
Foreground is the nonpositive region. These formulas specify implicit
shapes, not necessarily exact Euclidean signed distances everywhere.
Translation is uniform in $[-0.08,0.08]^2$, angle jitter in
$[-0.12,0.12]$ radians, and each axis scale in $[0.9,1.1]$.
The restricted rotation range preserves the square/diamond distinction.

Each class contributes 80 origins. Within class, seeds 0--49 train
readers, 50--59 validate them, and 60--79 test them; seeds 60--69 define
the language-model subset. A disjoint two-origin-per-class pilot selects
width 32 and 6,000 fitting updates by reconstruction quality.
Targets are evaluated on a $33\times33$ grid. Incoming weights and biases
initialize uniformly in $[-1/\sqrt2,1/\sqrt2]$; outgoing weights and bias
use $[-1/\sqrt h,1/\sqrt h]$. Adam \cite{kingma2015adam} uses learning rate
0.015, default moments, and 256 sampled grid coordinates per update.
Parameters and optimizer states are independent across origins.

\subsection{Digit fields}
For each MNIST class \cite{lecun1998mnist}, the first 100 training-set
images are averaged in $2\times2$ blocks to obtain $14\times14$ targets.
A width-64 ReLU field fits $0.5-I(x)$ for 6,000 updates, using learning
rate 0.012 and 96 sampled coordinates per update. Ordered images within
each class define 60/20/20 training/validation/test splits. PSNR uses
unclipped reconstructions and unit signal range. Clipping is used only
for illustrations. ReLU training uses FP32 PyTorch \cite{paszke2019pytorch};
stored parameters and the original orbit evaluations use FP64.

\section{ReLU Readers and Complete Results}\label{app:readers}
\subsection{Features and training}
Flat ridge concatenates all parameters. Stats ridge computes mean,
standard deviation, extrema, mean absolute value, three quantiles
(10/50/90\%), and root mean square for each of $W,b,v$, then appends $c$:
28 features in total. Norm+Stats first normalizes incoming affine rows.
Canon+Flat normalizes, sorts complete tuples, and concatenates. Probe
ridge instead queries the field on a $9\times9$ grid. This 81-query
control establishes accessible functional information.

Ridge readers fit one-hot targets with an unregularized intercept.
Training data determine feature means and standard deviations, with a
$10^{-5}$ floor. Validation selects regularization from
$\{10^{-3},10^{-2},10^{-1},1,10,10^2,10^3\}$, retaining the first maximum.

The set reader applies two width-64 linear/ReLU layers to each neuron
tuple, averages neuron embeddings, appends the standardized output bias,
and applies a width-64 readout. Parameter counts are 9,094 for shapes
and 9,354 for digits. Neuron standardization uses training data with a
$10^{-4}$ floor. Adam uses learning rate 0.002, decay $10^{-4}$, batch
size 64, and 2,000 updates. Validation every 100 updates selects the
first maximum; predictions average the selected logits of seeds 11,
22, and 33.

Set+Aug samples identity or a fresh joint transformation with equal
probability and selects on mean clean/transformed validation accuracy.
Other set readers select on clean validation. Normalized and canonical
features use the same fixed preprocessing in training and evaluation.
Permutation invariance of set aggregation is an arithmetic property;
finite-precision score differences are checked separately.

\begin{table*}[t]
\centering\small
\caption{\textbf{Shape readers: accuracy and agreement separate.} Percent on 120 test origins; each joint transformation is paired with its original. Dis. is prediction disagreement and Both is both-correct rate. Probe ridge executes the field; all other rows read weights. Confidence intervals are in the supplement.}\label{tab:shapes}
\begin{tabular}{lrrrrrr}
\toprule
Reader & Raw & Perm. & Scale & Joint & Dis. & Both \\
\midrule
Flat ridge & 32.5 & 38.3 & 15.8 & 20.0 & 75.0 & 10.8 \\
Stats ridge & 96.7 & 96.7 & 21.7 & 24.2 & 75.0 & 24.2 \\
Norm+Stats & 97.5 & 97.5 & 97.5 & 97.5 & 0.0 & 97.5 \\
Canon+Flat & 71.7 & 71.7 & 71.7 & 71.7 & 0.0 & 71.7 \\
Probe ridge (81 queries) & 100.0 & 100.0 & 100.0 & 100.0 & 0.0 & 100.0 \\
Set & 96.7 & 96.7 & 26.7 & 20.0 & 80.8 & 19.2 \\
Set+Aug & 100.0 & 100.0 & 95.8 & 95.0 & 5.0 & 95.0 \\
Norm+Set & 95.0 & 95.0 & 95.0 & 95.0 & 0.0 & 95.0 \\
\bottomrule
\end{tabular}
\end{table*}

\begin{table}[t]
\centering\small
\caption{\textbf{MNIST image fields.} Percent on 200 held-out images. Same structural controls, ten classes and width 64; language models are not evaluated on this dataset.}\label{tab:digits}
\begin{tabular}{lrrr}
\toprule
Reader & Raw & Joint & Dis. \\
\midrule
Flat ridge & 13.5 & 7.0 & 89.5 \\
Stats ridge & 33.0 & 11.5 & 89.0 \\
Norm+Stats & 33.5 & 33.5 & 0.0 \\
Canon+Flat & 33.5 & 33.5 & 0.0 \\
Probe ridge & 74.0 & 74.0 & 0.0 \\
Set & 68.5 & 20.0 & 83.5 \\
Set+Aug & 66.0 & 55.5 & 38.0 \\
Norm+Set & 69.0 & 69.0 & 0.0 \\
\bottomrule
\end{tabular}
\end{table}

\begin{table*}[t]
\centering\small
\caption{Complete shape reader intervals and paired flip counts. Accuracy intervals are Wilson; change intervals are stratified origin bootstrap, in percentage points.}\label{tab:shapeintervals}
\begin{tabular}{lrrrrr}
\toprule
Reader & Raw [95\% CI] & Joint [95\% CI] & Change [95\% CI] & C$\to$W & W$\to$C \\
\midrule
Flat ridge & 32.5 [24.8, 41.3] & 20.0 [13.8, 28.0] & -12.5 [-22.5, -3.3] & 26 & 11 \\
Stats ridge & 96.7 [91.7, 98.7] & 24.2 [17.4, 32.6] & -72.5 [-78.3, -66.7] & 87 & 0 \\
Norm+Stats & 97.5 [92.9, 99.1] & 97.5 [92.9, 99.1] & 0.0 [0.0, 0.0] & 0 & 0 \\
Canon+Flat & 71.7 [63.0, 79.0] & 71.7 [63.0, 79.0] & 0.0 [0.0, 0.0] & 0 & 0 \\
Probe ridge (81 queries) & 100.0 [96.9, 100.0] & 100.0 [96.9, 100.0] & 0.0 [0.0, 0.0] & 0 & 0 \\
Set & 96.7 [91.7, 98.7] & 20.0 [13.8, 28.0] & -76.7 [-80.8, -71.7] & 93 & 1 \\
Set+Aug & 100.0 [96.9, 100.0] & 95.0 [89.5, 97.7] & -5.0 [-9.2, -1.7] & 6 & 0 \\
Norm+Set & 95.0 [89.5, 97.7] & 95.0 [89.5, 97.7] & 0.0 [0.0, 0.0] & 0 & 0 \\
\bottomrule
\end{tabular}
\end{table*}

\begin{table*}[t]
\centering\small
\caption{Complete digit reader intervals and paired flip counts. Accuracy intervals are Wilson; change intervals are stratified origin bootstrap, in percentage points.}\label{tab:digitintervals}
\begin{tabular}{lrrrrr}
\toprule
Reader & Raw [95\% CI] & Joint [95\% CI] & Change [95\% CI] & C$\to$W & W$\to$C \\
\midrule
Flat ridge & 13.5 [9.4, 18.9] & 7.0 [4.2, 11.4] & -6.5 [-12.0, -1.0] & 25 & 12 \\
Stats ridge & 33.0 [26.9, 39.8] & 11.5 [7.8, 16.7] & -21.5 [-27.5, -16.0] & 56 & 13 \\
Norm+Stats & 33.5 [27.3, 40.3] & 33.5 [27.3, 40.3] & 0.0 [0.0, 0.0] & 0 & 0 \\
Canon+Flat & 33.5 [27.3, 40.3] & 33.5 [27.3, 40.3] & 0.0 [0.0, 0.0] & 0 & 0 \\
Probe ridge (81 queries) & 74.0 [67.5, 79.6] & 74.0 [67.5, 79.6] & 0.0 [0.0, 0.0] & 0 & 0 \\
Set & 68.5 [61.8, 74.5] & 20.0 [15.0, 26.1] & -48.5 [-56.0, -40.5] & 112 & 15 \\
Set+Aug & 66.0 [59.2, 72.2] & 55.5 [48.6, 62.2] & -10.5 [-17.0, -4.0] & 38 & 17 \\
Norm+Set & 69.0 [62.3, 75.0] & 69.0 [62.3, 75.0] & 0.0 [0.0, 0.0] & 0 & 0 \\
\bottomrule
\end{tabular}
\end{table*}

\subsection{Intervals and numerical meaning}
Wilson intervals describe rates over origins. Paired changes use 2,000
class-stratified percentile bootstrap draws, carrying all paired
conditions together. These intervals condition on the trained ensemble
and sampled population. A collapsed interval for identical observed
changes describes the empirical resampling distribution; it does not
certify zero population risk. Complete flip counts distinguish harmful,
helpful, and incorrect-to-incorrect changes.

\section{Numerical Fidelity}
\label{app:numerical}
The ReLU stress design samples integer or continuous neuron-wise
exponents from $[-a,a]$, for $a\in\{4,8,16\}$, with four repetitions.
Each pair is executed on 4,096 shared fresh coordinates at 2, 3, 4, 6,
and 8 significant digits. This gives 14,400 shape and 24,000 digit
precision records. The grid is not used for reader training.
The audit distinguishes full-precision equivalence, differences between
separately rounded representations, and each rounded representation's
error relative to its original field.

\begin{table}[!htbp]
\centering\small
\caption{Post hoc reconstruction-stratified set-reader accuracy (percent). Quartiles are formed within each class, low to high quality. Each shape quartile has 30 origins and each MNIST quartile 50. The coordinate failure persists in well-fitted fields.}\label{tab:quality}
\begin{tabular}{llrr}
\toprule
Dataset & Quartile & Raw & Joint \\
\midrule
Shapes & Q1 & 93.3 & 23.3 \\
Shapes & Q2 & 96.7 & 16.7 \\
Shapes & Q3 & 96.7 & 20.0 \\
Shapes & Q4 & 100.0 & 20.0 \\
MNIST & Q1 & 66.0 & 20.0 \\
MNIST & Q2 & 72.0 & 18.0 \\
MNIST & Q3 & 66.0 & 20.0 \\
MNIST & Q4 & 70.0 & 22.0 \\
\bottomrule
\end{tabular}
\par\medskip
\centering\small
\caption{\textbf{Uniform six-digit output-error bounds.} Maximum conservative bounds over test origins, valid throughout $[-1,1]^2$. The pair compares rounded raw/joint fields; remaining rows compare with the original.}\label{tab:uniform}
\begin{tabular}{lrr}\toprule
Representation & Shapes & MNIST \\ \midrule
raw & $3.93\times10^{-5}$ & $2.73\times10^{-4}$ \\
joint & $4.09\times10^{-5}$ & $2.40\times10^{-4}$ \\
canonical & $3.34\times10^{-5}$ & $1.76\times10^{-4}$ \\
raw\_joint\_pair & $7.45\times10^{-5}$ & $4.82\times10^{-4}$ \\
\bottomrule\end{tabular}
\end{table}

The primary six-digit strings additionally admit a uniform bound.
For a reference neuron, its ReLU value is at most
$M_j=\max(0,b_j+\|w_j\|_1)$ on $[-1,1]^2$. The discrepancy in its
activation is at most
$\delta_j=\|w_j-\widehat w_j\|_1+|b_j-\widehat b_j|$.
Adding and subtracting
$\widehat v_j\relu(w_j^\top x+b_j)$ gives
\[
 |f(x)-\widehat f(x)|\leq |c-\widehat c|
 +\sum_j\bigl(|v_j-\widehat v_j|M_j+|\widehat v_j|\delta_j\bigr).
\]
Stored binary weights and displayed decimals are converted to exact
rational numbers. Power-of-two transformations are verified exactly.
For canonical inputs, the computed normalization factor is treated as
an exact positive rational scale, defining a function-equivalent
reference before rounding. The bound includes normalization error,
uses the serializer's sorting permutation, and maps zero rows to zero.
Reported bounds are rounded upward.

This certificate concerns the real-arithmetic functions defined by the
printed numbers, not arbitrary floating-point evaluation kernels.
Foreground sign is preserved wherever the reference magnitude exceeds
the bound; no sign guarantee follows at the zero-level set. Post hoc
quality quartiles show that the set-reader failure persists among the
best fits. Canonical strings coincide for the primary equivalent pairs,
providing an input-equality control for deterministic readers.

\section{Natural-Image Orbit Study}\label{app:orbit_extension}
\subsection{Populations and fitting}
The 1,600-image population has 1,000 training, 200 validation, and 400
balanced official-test images. It is disjoint from calibration and the
600-image source pilot. Three fitting bundles each supply a training
source S0 and a held-out source S2. Each source shares an initialization
across images, while bundle/source initializations differ. All 9,600
fields use architecture $2$--$64$--$64$--$3$, sine frequency 8, and a
linear output. Adam fits standardized CIFAR10 RGB values on the full
$32\times32$ grid for 1,000 updates at learning rate $10^{-3}$.
A separate calibration fixes architecture and fitting duration.

Mean test reconstruction PSNR is 31.44/32.60/31.61\,dB for the three S0
bundles and 31.89/32.12/32.48\,dB for S2. The lowest retained test PSNR
is 27.23\,dB. No low-quality fit is removed.

\subsection{Reader implementations}
HiddenProbe reads both hidden-layer response profiles and the output,
using the upstream set-transformer head with width 120, eight attention
heads, two encoder layers, and dropout 0.1. ProbeGen uses its two-layer
linear generator, 128 probes, six mixer layers of width 318, and a
$3$--$256$--$4$ per-probe network. Both coordinate generators use the
same initialization and a tanh map into $[-1,1]^2$. This domain
restriction and HiddenProbe's EMA decay of 0.99 are small-data adaptations.

MPGMN uses a four-layer width-128 forward matrix-product graph reader,
mean aggregation, fan-in-rescaled edges, and layerwise readout. It is
configured for permutation symmetry; ReLU scale normalization is not
applied to sine networks. Grid64 and Grid256 query $8\times8$ and
$16\times16$ grids and use two width-128 layers. WeightStats standardizes
42 tensor summaries using training data. DenseImage uses convolution
widths 32/32/64/128, group normalization, pooling, and a width-128
readout. The same selected image classifier reads reconstructed test
images under every orbit condition.

HiddenPhase replaces preactivations by their wrapped phase.
HiddenAugment trains with independent signed one-period bias shifts.
HiddenGram replaces each 263-dimensional profile feature by
\[
 \Phi(h)=(h_1h,h_2h,\nu(h)),\qquad h\in\R^{128}.
\]
The seven entries of $\nu$ are squared mean, standard deviation,
maximum/minimum absolute value, mean absolute value, mean square,
and Euclidean norm. Each is even under $h\mapsto-h$. Probe positions,
head, parameter count, optimizer, EMA, and selection schedule match
HiddenProbe. The anchors are the first two fixed probe positions.

Whole-period shifts leave $h$ unchanged; compensated negations change
it only by a global sign. Since $(-h)_a(-h)=h_a h$, both actions leave
$\Phi$ unchanged, and set aggregation handles neuron permutations.
If either anchor is nonzero, its Gram row identifies the profile up
to sign. No completeness is claimed when both anchors vanish or for
all functionally equivalent networks.

The SIREN-GAP comparison uses the published phasor-graded W12 reader
\cite{sirengap2026software}. The adapter transposes kernels and absorbs
$\omega$ into hidden kernels and biases. Train-only feature scaling,
RGB output channels, and graded processing follow the implementation.
Its width-selection rule targets approximately 1.87 million parameters;
it is not capacity-matched to HiddenProbe. AdamW uses learning rate
$10^{-3}$, decay 0.01, and a cosine schedule.

\subsection{Training and uncertainty}
Nine original methods across three reader seeds and three fitting bundles
supply 81 states; the two auxiliary controls add nine states each.
Every state receives 240 epochs, 240,000 image presentations, and
7,680 updates. All 99 validation selections precede test access.
The matched HiddenProbe controls share the same training protocol;
other architectures retain method-specific optimization. The complete
capacity and selection ranges are reported in \cref{tab:orbit_cost}.

The primary contrast is HiddenProbe Phase 1 minus raw accuracy on S0;
the analogous HiddenPhase contrast is co-primary. Exact sign
randomization treats each image's change vector as one exchangeable
cluster, with Holm adjustment across the two tests. Phase 3, sign,
other methods, and S2 transfer are descriptive comparisons.
The 10,000-draw class-stratified image bootstrap carries every condition,
seed, and bundle with its image. Nine reader outcomes for one image
remain one origin cluster.

\subsection{Numerical checks and complete results}
Transformations are deterministic functions of image identity and do
not depend on batch position. Checks cover the 1,024 image coordinates,
1,024 independent coordinates, and learned probe banks. Ideal FP64
actions and separately rounded FP32 actions are compared with the
original field. Their maximum observed errors are
$1.75\times10^{-14}$ and $1.12\times10^{-5}$ in standardized RGB units.
These finite-query checks complement the exact-arithmetic identities.

\begin{table*}[t]
\centering\footnotesize\setlength{\tabcolsep}{4pt}
\caption{\textbf{S0: fitted training source.} Accuracy and phase-one change have 95\% original-image intervals. Remaining entries are means over nine states. $D$ is paired prediction disagreement (\%). All permutation disagreements are zero.}\label{tab:orbit_s0}
\begin{tabular}{lccrrrr}
\toprule
Reader & Raw accuracy (\%) & Phase-one change (pp) & Phase-one $D$ & Phase-three acc. & Sign acc. & Sign $D$\\
\midrule
HiddenProbe & 30.0 [27.1, 33.0] & -8.1 [-10.4, -6.0] & 61.6 & 14.2 & 29.4 & 27.0\\
HiddenPhase & 29.3 [26.4, 32.2] & 0.0 [0.0, 0.0] & 0.0 & 29.3 & 28.6 & 19.1\\
HiddenAugment & 29.4 [26.5, 32.3] & 0.8 [-0.1, 1.8] & 18.0 & 27.8 & 29.3 & 13.2\\
HiddenGram & 28.7 [25.9, 31.6] & 0.0 [0.0, 0.0] & 0.0 & 28.7 & 28.7 & 0.0\\
PhasorW12 & 24.5 [22.5, 26.6] & 0.0 [0.0, 0.0] & 0.0 & 24.5 & 24.5 & 0.0\\
ProbeGen & 29.2 [26.3, 32.1] & 0.0 [0.0, 0.0] & 0.0 & 29.2 & 29.2 & 0.0\\
MPGMN & 25.2 [23.0, 27.4] & -12.3 [-14.8, -9.9] & 81.6 & 10.8 & 17.4 & 74.2\\
Grid64 & 34.9 [31.2, 38.6] & 0.0 [0.0, 0.0] & 0.0 & 34.9 & 34.9 & 0.0\\
Grid256 & 33.1 [29.4, 36.8] & 0.0 [0.0, 0.0] & 0.0 & 33.1 & 33.1 & 0.0\\
WeightStats & 23.8 [21.5, 26.1] & -13.0 [-15.3, -10.7] & 87.5 & 10.2 & 12.9 & 84.5\\
DenseImage & 44.1 [40.8, 47.6] & 0.0 [0.0, 0.0] & 0.0 & 44.1 & 44.1 & 0.0\\
\bottomrule
\end{tabular}
\end{table*}

\begin{table*}[t]
\centering\footnotesize\setlength{\tabcolsep}{4pt}
\caption{\textbf{S2: unseen fitting source.} Accuracy and phase-one change have 95\% original-image intervals. Remaining entries are means over nine states. $D$ is paired prediction disagreement (\%). All permutation disagreements are zero.}\label{tab:orbit_s2}
\begin{tabular}{lccrrrr}
\toprule
Reader & Raw accuracy (\%) & Phase-one change (pp) & Phase-one $D$ & Phase-three acc. & Sign acc. & Sign $D$\\
\midrule
HiddenProbe & 30.1 [27.2, 33.1] & -9.1 [-11.1, -7.0] & 60.2 & 13.9 & 30.0 & 24.7\\
HiddenPhase & 29.4 [26.6, 32.4] & 0.0 [0.0, 0.0] & 0.0 & 29.4 & 29.3 & 18.9\\
HiddenAugment & 30.0 [27.1, 32.9] & 0.4 [-0.6, 1.4] & 17.5 & 28.2 & 28.8 & 12.5\\
HiddenGram & 29.7 [26.8, 32.6] & 0.0 [0.0, 0.0] & 0.0 & 29.7 & 29.7 & 0.0\\
PhasorW12 & 20.3 [18.7, 22.0] & 0.0 [0.0, 0.0] & 0.0 & 20.3 & 20.3 & 0.0\\
ProbeGen & 29.2 [26.4, 32.1] & 0.0 [0.0, 0.0] & 0.0 & 29.2 & 29.2 & 0.0\\
MPGMN & 17.1 [15.6, 18.6] & -6.1 [-7.9, -4.2] & 76.9 & 10.8 & 17.3 & 70.4\\
Grid64 & 34.8 [31.2, 38.6] & 0.0 [0.0, 0.0] & 0.0 & 34.8 & 34.8 & 0.0\\
Grid256 & 33.0 [29.3, 36.8] & 0.0 [0.0, 0.0] & 0.0 & 33.0 & 33.0 & 0.0\\
WeightStats & 12.9 [11.8, 14.0] & -2.2 [-3.3, -1.2] & 78.0 & 10.2 & 11.8 & 52.1\\
DenseImage & 44.0 [40.7, 47.4] & 0.0 [0.0, 0.0] & 0.0 & 44.0 & 44.0 & 0.0\\
\bottomrule
\end{tabular}
\end{table*}

\begin{table*}[t]
\centering\footnotesize
\caption{Per-bundle S0 accuracy (\%). Each cell averages three reader seeds; these are three fitting bundles, not additional independent image populations. Complete per-seed values are included in the aggregate records.}\label{tab:orbit_bundles}
\begin{tabular}{lrrrrrr}
\toprule
& \multicolumn{2}{c}{Bundle 0} & \multicolumn{2}{c}{Bundle 1} & \multicolumn{2}{c}{Bundle 2}\\
Reader & Raw & Phase one & Raw & Phase one & Raw & Phase one\\
\midrule
HiddenProbe & 26.8 & 17.1 & 31.4 & 25.2 & 31.8 & 23.3\\
HiddenPhase & 28.0 & 28.0 & 27.8 & 27.8 & 32.0 & 32.0\\
HiddenAugment & 27.8 & 28.8 & 30.1 & 31.4 & 30.2 & 30.4\\
HiddenGram & 28.8 & 28.8 & 29.3 & 29.3 & 27.9 & 27.9\\
PhasorW12 & 22.9 & 22.9 & 24.1 & 24.1 & 26.5 & 26.5\\
ProbeGen & 29.6 & 29.6 & 29.8 & 29.8 & 28.1 & 28.1\\
MPGMN & 24.3 & 13.4 & 24.3 & 13.8 & 27.1 & 11.4\\
Grid64 & 35.2 & 35.2 & 34.6 & 34.6 & 34.8 & 34.8\\
Grid256 & 34.2 & 34.2 & 32.2 & 32.2 & 32.7 & 32.7\\
WeightStats & 21.2 & 10.0 & 25.6 & 10.0 & 24.4 & 12.3\\
DenseImage & 44.4 & 44.4 & 44.0 & 44.0 & 44.0 & 44.0\\
\bottomrule
\end{tabular}
\end{table*}

\begin{table}[t]
\centering\small
\setlength{\tabcolsep}{5pt}
\caption{\textbf{Reader capacity and model selection.} Parameter counts are per state; selected-epoch ranges span nine states per architecture in the natural-image orbit study. The source-transfer pilot uses the same seven corresponding reader architectures.}\label{tab:orbit_cost}
\begin{tabular}{lrr}
\toprule
Reader & Parameters & Selected epochs \\
\midrule
HiddenProbe & 447,908 & 30--200 \\
HiddenPhase & 447,908 & 40--220 \\
HiddenAugment & 447,908 & 20--220 \\
HiddenGram & 447,908 & 20--240 \\
PhasorW12 & 1,875,270 & 20--180 \\
ProbeGen & 579,362 & 10--150 \\
MPGMN & 480,394 & 40--220 \\
Grid64 & 42,506 & 10--90 \\
Grid256 & 116,234 & 10--140 \\
WeightStats & 23,306 & 10--90 \\
DenseImage & 366,250 & 10--220 \\
\bottomrule
\end{tabular}
\end{table}

The complete test matrix contains 396,000 predictions. HiddenGram and
the phasor reader have unchanged class predictions under every tested
action, but their maximum S0 Phase 3 logit drifts are
$3.17\times10^{-4}$ and $2.38\times10^{-4}$, respectively. The
floating-point implementations are therefore not claimed to be bitwise
invariant. HiddenPhase's zero phase-accuracy change gives exact $p=1$;
the structural construction and prediction checks, rather than that
nonrejection, support its phase-invariance interpretation.

On the unseen S2 source, HiddenProbe's raw accuracy changes by
$+0.06$ points $[-1.31,1.44]$, while MPGMN loses 8.14 points.
Exact-orbit and refitting outcomes need not coincide. The Gram--phasor
raw difference is $4.19$ points $[1.47,6.94]$, but their capacity and
optimization differ; this is a protocol-specific comparison, not an
architecture-only effect or a ranking on upstream benchmarks.

\section{Feature Location, Canonical Training, and Refitting}
\label{app:diagnostics}
These retrospective diagnostics use retained fields and reader states
from the earlier experiments. They test the scope of the main
mechanistic interpretation and are not additional prospective primary
comparisons.

\subsection{Feature-location comparisons}
The CIFAR10 study uses 400 training, 100 validation, and 100 audit images
from the source pilot. Features are final RGB outputs, first-layer
post-sine activations, first-layer preactivations, or tensor statistics.
The first three use a $12\times12$ probe grid. Hidden profiles are
projected through fixed, data-independent Gaussian matrices to 432
features; outputs have 432 features and statistics have 24. Each reader
has two width-256 hidden layers, AdamW learning rate 0.002, decay
$10^{-4}$, batch size 32, and 60 epochs. Training-only standardization
precedes the head; the first maximum validation accuracy selects the
state. Five initialization seeds are reported.

The 16-entry transformation design contains the original and repeated
instances of four fixed per-image actions: permutation, Phase 1,
Phase 3, and negation. The action generator reuses its per-image seed
for each occurrence of an action. These entries define a weighted
average over the fixed actions, not 16 independent orbit draws.
The supplementary eight-entry variant uses the same fixed-action
structure. This distinction is relevant to interpreting an averaging
budget, but does not change the retained measured accuracies.

The MNIST comparison uses a $13\times13$ grid, 169 output features,
507 projected hidden features, or nine weight statistics. The head
width, optimizer, and learning rate match the CIFAR10 comparison;
training uses batch size 64 and 80 epochs. Its fifteen transformed
entries use distinct seeded permutation/rescaling draws, some of which
are identity actions. The reported five seeds share the same 200 test
images.

These comparisons control head depth and hidden width, training
schedule, split, and seed choices. They do not hold input-layer parameter
count or feature information fixed. Hidden profiles have neuron
coordinates and may omit the output weights; fixed projection also
changes their information. Thus the results establish sensitivity of
the specified readouts, not a pure causal effect of feature location
independent of capacity and representation.

\subsection{Canonical inputs during training and evaluation}
The separate six-seed MNIST experiment uses unprojected hidden profiles.
It compares original inputs with normalization and lexicographic
sorting applied consistently in training, validation, and evaluation.
The section sorts $(\bar b,\bar w,\bar v)$; the earlier canonical flat
reader sorts $(\bar w,\bar b,\bar v)$. Either fixed total order defines
a representative for the tested action in exact arithmetic.
Appending outgoing weights tests one source of information loss, but
still omits the output bias from those hidden-profile features.

\begin{table*}[t]
\centering\small
\setlength{\tabcolsep}{5pt}
\caption{\textbf{Canonical training on MNIST.} Six-seed means on the same 200 test images. Hidden profiles are unprojected here; this is separate from the main-paper feature-location study. Small nonzero $G_o$ values remain after finite-precision normalization and sorting.}\label{tab:canonical_readouts}
\begin{tabular*}{\textwidth}{@{\extracolsep{\fill}}llrrr@{}}
\toprule
Features & Train/test inputs & Accuracy (\%) & Averaging change (pp) & $G_o$ \\
\midrule
Field outputs & Original & 81.25 & +0.00 & $<10^{-9}$ \\
Hidden activations & Original & 15.17 & -0.67 & 0.87217 \\
Hidden + outgoing & Original & 14.17 & -0.50 & 0.87735 \\
Hidden activations & Canonical & 38.50 & +0.00 & 0.00050 \\
Hidden + outgoing & Canonical & 40.08 & +0.00 & 0.00054 \\
Preactivation + outgoing & Canonical & 29.75 & +0.00 & 0.00119 \\
Weight statistics & Original & 31.25 & -18.42 & 0.86194 \\
Weight statistics & Canonical & 34.50 & +0.00 & 0.00008 \\
\bottomrule
\end{tabular*}
\end{table*}

\begin{table}[t]
\centering\small
\setlength{\tabcolsep}{5pt}
\caption{\textbf{Feature-location results on the larger held-out split.} 1,600 CIFAR10 origins (1,000 train / 200 validation / 400 test), fitted twice independently. Between-fit PSNR is 34.8 dB and fit PSNR is 31.8 dB, both in RGB units. Accuracy averages five seeds on the 400 test origins, never used for selection. Head depth, hidden width, optimiser, budget, and split are shared; unequal input dimensions mean parameter counts and information differ.}\label{tab:main_scale}
\begin{tabular}{lrrr}
\toprule
Reader features & $G_o$ & Test (\%) & Averaged (\%) \\
\midrule
Field outputs & $<10^{-9}$ & 33.3 & 33.3 \\
Hidden activations & 0.778 & 8.4 & 8.2 \\
Hidden preactivations & 0.858 & 11.6 & 10.1 \\
Weight statistics & 0.835 & 18.1 & 11.5 \\
\bottomrule
\end{tabular}
\end{table}

On a larger independently fitted population (\cref{tab:main_scale}) the same ordering holds: the output reader reaches 33.3\% on 400 held-out test images while hidden and weight features stay between 8.4\% and 18.1\%, and each hidden-feature point estimate decreases under averaging. The weight-statistic change is $-6.60$ points with a descriptive origin-bootstrap interval $[-9.25,-4.05]$; the hidden-activation and preactivation intervals include zero. Seeds are kept together within each of the 400 origins. The field-output reader is unchanged to the last digit because no transformation in the audited group alters the rendered function.

The canonical hidden reader reaches 38.5\% mean accuracy versus 15.2\%
for original hidden profiles. Its mean orbit-variance share is
$4.98\times10^{-4}$, not exactly zero; normalization and sorting have
finite-precision sensitivity. Averaging leaves the observed class
accuracy unchanged. A theoretical invariant representation, stable
class decisions, and identical numerical scores are distinct claims.
No significance test treats the six predictions of each image as six
independent observations.

\paragraph{Complete-input follow-up.}
A separate CPU replay uses the same 600/200/200 split, six seeds
(11, 22, 33, 44, 55, 66), $13\times13$ queries, two width-256 hidden
layers, 80 epochs, batch size 64, and first-best validation selection.
Five conditions give 30 trained states (\cref{tab:complete_inputs}).
The complete input concatenates every hidden activation, $v$, and $c$;
no random projection is applied. The fixed-composition control extracts
these values from that concatenated vector and computes
$\sum_jv_jh_j(x)+c$. Its query values equal direct evaluation of the
canonical field bit for bit. Normalizing and sorting the original field
can still introduce finite-precision differences.

\begin{table*}[t]
\centering\small
\setlength{\tabcolsep}{5pt}
\caption{\textbf{Complete-input control on MNIST.} Six-seed means on 200 independent test origins; canonical hidden inputs are used in training and evaluation. $v,c$ are outgoing weights and bias. Fixed composition reconstructs the queried field outputs from the complete hidden vector. Heads share depth and hidden width; input sizes and parameter counts differ.}\label{tab:complete_inputs}
\begin{tabular*}{\textwidth}{@{\extracolsep{\fill}}lrrrr@{}}
\toprule
Features & Input size & Parameters & Accuracy (\%) & $G_o$ \\
\midrule
Direct outputs (original) & 169 & 111,882 & 81.33 & $<10^{-9}$ \\
Canonical hidden & 10816 & 2,837,514 & 37.75 & 0.00035 \\
Hidden + $v$ & 10880 & 2,853,898 & 39.92 & 0.00060 \\
Hidden + $v,c$ & 10881 & 2,854,154 & 38.67 & 0.00076 \\
Fixed output composition & 169 & 111,882 & 81.50 & $<10^{-9}$ \\
\bottomrule
\end{tabular*}
\end{table*}

The complete-input reader reaches 38.7\%, while fixed composition reaches
81.5\%, a 42.83-point difference with descriptive paired interval
$[36.58,48.83]$. Direct outputs reach 81.3\%; the composition-minus-output
interval is $[-0.67,1.00]$ points. Intervals use 10,000 class-stratified
origin resamples, carrying all six seeds together, and condition on the
trained states. These are retrospective comparisons. The CPU replay is
reported separately from the earlier GPU run, rather than substituting
new values into its historical seed means. The much larger input layer
of the hidden readers is not parameter-matched to the output readers;
these controls establish recoverable information, not a capacity-isolated
cause of the learning gap.

\begin{table*}[t]
\centering\small
\setlength{\tabcolsep}{5pt}
\caption{\textbf{Matching training and evaluation coordinates on shapes.} Four independent seed offsets; accuracy is mean $\pm$ population standard deviation across offsets. Disagreement is the mean paired prediction-change rate. Augmentation is present only for Set+Aug; the generic source protocol label does not override the reader configuration.}\label{tab:canonical_shapes}
\begin{tabular*}{\textwidth}{@{\extracolsep{\fill}}llrr@{}}
\toprule
Reader & Training / evaluation inputs & Accuracy (\%) & Disagree (\%) \\
\midrule
Stats ridge & Original / original & 96.67 $\pm$ 0.00 & 74.38 \\
Stats ridge & Canonical / canonical & 97.50 $\pm$ 0.00 & 0.00 \\
Norm+Stats & Original / original & 97.50 $\pm$ 0.00 & 0.00 \\
Norm+Stats & Canonical / canonical & 97.50 $\pm$ 0.00 & 0.00 \\
Set & Original / original & 96.25 $\pm$ 0.42 & 80.42 \\
Set & Canonical / canonical & 95.62 $\pm$ 0.91 & 0.00 \\
Norm+Set & Original / original & 95.83 $\pm$ 0.59 & 0.00 \\
Norm+Set & Canonical / canonical & 95.62 $\pm$ 0.91 & 0.00 \\
Set+Aug & Original / original & 99.79 $\pm$ 0.36 & 6.46 \\
Set+Aug & Canonical / canonical & 98.96 $\pm$ 0.36 & 0.00 \\
Set+Aug & Augmented original / canonical & 96.67 $\pm$ 1.77 & 0.00 \\
\bottomrule
\end{tabular*}
\end{table*}

The four-offset shape study gives Stats ridge 96.67\% original and
97.50\% canonical accuracy, while the set reader changes from 96.25\%
to 95.63\%. Set+Aug reaches 99.79\% on original inputs and 98.96\%
with canonical training/evaluation. It reaches 96.67\% when trained
with augmentation on original inputs and evaluated canonically.
These measurements do not establish that canonicalization outperforms
augmentation. Rows labeled generically as an augmentation-to-section
protocol in the source output are interpreted using their actual reader
configuration: an unaugmented set reader remains unaugmented.

\subsection{Orbit variation and source variation}
For origin $i$, source $s$, and transformation $g$, let $P_{isg}$ be the
predicted probability vector. Put $\bar P_{is}=\E_gP_{isg}$ and
$\bar P_i=\E_s\bar P_{is}$. The crossed design measures
\begin{align*}
 V_g&=\E_{i,s,g}\|P_{isg}-\bar P_{is}\|^2,\\
 V_f&=\E_{i,s}\|\bar P_{is}-\bar P_i\|^2,\\
 G_f&=V_g/(V_g+V_f).
\end{align*}
The 42 states cover seven readers, three seeds, and two training regimes.
They are evaluated on 100 audit images, four fitting sources, and eight
transformation entries. Probabilities, not logits, enter the statistic.
The retained implementation scales both variance components by the
same factor; this cancels in $G_f$. Its separately stored object
component has a different normalization and is not combined with those
raw components here.

\begin{table}[t]
\centering\small
\setlength{\tabcolsep}{5pt}
\caption{\textbf{Orbit variation relative to fitting-source variation.} Each row spans three seeds in two training regimes (six states), evaluated on 100 origins and four fitting sources. $G_f$ uses between-fit variation in its denominator; it is not $G_o$. The averaging change uses this eight-entry design.}\label{tab:orbit_fit_share}
\begin{tabular}{lrr}
\toprule
Reader & $G_f$ range & Accuracy change (pp) \\
\midrule
HiddenProbe & 0.925--0.984 & -1.50 to +2.25 \\
MPGMN & 0.848--0.893 & -4.00 to +2.25 \\
WeightStats & 0.926--0.985 & -7.25 to -1.75 \\
ProbeGen & $<10^{-9}$ & +0.00 to +0.00 \\
Grid64 & $<10^{-9}$ & +0.00 to +0.00 \\
Grid256 & $<10^{-9}$ & +0.00 to +0.00 \\
DenseImage & $<10^{-9}$ & +0.00 to +0.00 \\
\bottomrule
\end{tabular}
\end{table}

All 24 output-based states have $G_f<10^{-9}$. All 18 hidden/weight
states exceed 0.848, but some of those states gain accuracy from averaging
and others lose it. The observed separation describes these architectures
and this perturbation distribution. It does not validate a universal
cutoff, a predictor of the sign of accuracy change, or the assertion
that symmetry sensitivity cannot be learned away.

\subsection{Bandwidth and agreement between fits}
The bandwidth experiment fits 192 targets twice, using separate
initializations but the same SIREN architecture and 1,000-step schedule.
A radial Fourier mask retains frequencies up to a fraction of the
Nyquist frequency; the 1.00 condition uses the unfiltered image.
Reconstruction PSNR is always measured against the original image.
Between-fit PSNR measures agreement between the two reconstructed
images on the image grid. Both errors are converted from standardized
channels to unclipped $[0,1]$ RGB before squaring, with peak value one.
The 192-image experiment is replayed on CPU with the same seeds and
fitting recipe; its rendered predictions are retained for metric replay.

\begin{figure}[t]
\centering\includegraphics[width=\linewidth]{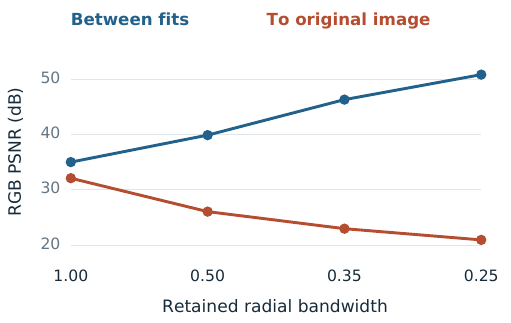}
\caption{\textbf{Smoothing trades image fidelity for fitting repeatability.}
Two fits of each of 192 images. Agreement between fits improves as
bandwidth decreases, while fidelity to the original image falls.
The horizontal positions denote the four tested settings.}
\label{fig:bandwidth}
\end{figure}
\begin{table}[t]
\centering\small
\setlength{\tabcolsep}{5pt}
\caption{\textbf{Repeatable fitting and recognition are distinct outcomes.} Fit is PSNR to the original image and Pair is between-fit PSNR, both in unclipped RGB units on 192 images (CPU replay). Accuracy uses only the 100 audit images, averaged over three reader seeds; the validation images used for selection are excluded. The original target is unfiltered at bandwidth 1.00.}\label{tab:bandwidth}
\begin{tabular}{rrrr}
\toprule
Band & Fit (dB) & Pair (dB) & Acc. (\%) \\
\midrule
1.00 & 32.09 & 35.00 & 27.0 \\
0.50 & 26.06 & 39.87 & 25.7 \\
0.35 & 22.98 & 46.29 & 26.7 \\
0.25 & 20.95 & 50.80 & 24.3 \\
\bottomrule
\end{tabular}
\end{table}

A separate three-seed reader experiment retains correctness for 100
validation and 100 audit images at each bandwidth. We recompute accuracy
on the 100 audit images only, excluding the validation images used for
checkpoint selection. It does not show a recognition gain at the strongest
filtering level. Because the target changes, the intervention supports
a tradeoff between repeatability and retained detail, not the claim
that high-frequency ambiguity explains all downstream error or that
bandwidth has no effect on recognition in general.

\section{Fitting-Source Transfer Pilot}\label{app:crossfit}
\subsection{Design and populations}
This pilot compares separate fits of the same CIFAR10 image. It therefore
tests source transfer, not exact function preservation. The class label
remains shared, while the fitted RGB field may change.
The balanced population contains 400 training, 100 validation, and
100 audit images, disjoint from 64 calibration images and the later
natural-image orbit population. All 600 images come from the official
training partition.

S0, S1, and S2 each use a shared initialization; I0 uses image-specific
initializations. The 2,400 fields have the architecture and fitting
schedule described for the orbit study. Calibration compares widths
32/64 and 1,000/5,000 updates, selecting the least compute proxy that
meets median PSNR at least 30\,dB and tenth-percentile PSNR at least
25\,dB for every source. All fields are retained.

Seven readers, three seeds, and two training regimes give 42 selected
states. One regime uses S0 only; the other uses equal S0/S1 presentations
without increasing the number of images or updates. S2 and I0 never
enter training or model selection. The dense CNN trains on original RGB;
its second regime is a training repeat rather than a change of source.

\subsection{Training and paired inference}
Readers use the adapters specified for the orbit study, with 240 epochs,
batch size 32, gradient norm cap one, and a 100-update warmup.
AdamW learning rates are $7\times10^{-4}$ for ProbeGen,
$5\times10^{-4}$ for MPGMN, and $10^{-3}$ for simple controls.
HiddenProbe uses $2\times10^{-4}$ with decay 0.1 for the head and
$7\times10^{-4}$ without decay for probe generation. ProbeGen has
zero decay; remaining controls use $10^{-4}$.
A validation-plateau scheduler has factor 0.7, patience five checks,
and floor $10^{-6}$. Validation after epoch one and every ten epochs
selects the first maximum.

All states are fixed before audit evaluation. Paired 10,000-draw
class-stratified bootstraps resample images with all sources and seeds.
FP32 training disables TF32; bitwise deterministic retraining is not
assumed. The two original-RGB CNN repetitions select slightly different
states, so their differences are not interpreted as source-balancing
benefits.

\begin{table*}[t]
\centering\small
\caption{Complete pilot source matrix (\%). Each cell is mean accuracy over reader seeds 11/22/33, with the three individual accuracies in parentheses. All conditions use the same 100 audit images. The Dense CNN second run repeats original-image training; its differences are not an effect of source-balanced inputs.}
\label{tab:crossfit-full}
\begin{tabular}{llrrrr}\toprule
Reader & Training & S0 & S1 & S2 & I0\\\midrule
HiddenProbe & S0 & 25.0 (31/23/21) & 25.3 (29/27/20) & 23.3 (30/20/20) & 24.3 (31/23/19)\\
HiddenProbe & S0/S1 & 22.3 (21/24/22) & 24.0 (23/27/22) & 21.7 (19/24/22) & 23.3 (24/23/23)\\
ProbeGen & S0 & 25.7 (27/27/23) & 26.0 (27/28/23) & 26.3 (27/29/23) & 26.0 (26/29/23)\\
ProbeGen & S0/S1 & 22.7 (24/20/24) & 23.3 (26/20/24) & 23.0 (25/20/24) & 22.7 (24/20/24)\\
MPGMN & S0 & 24.7 (34/23/17) & 12.7 (11/17/10) & 16.3 (19/14/16) & 18.0 (17/21/16)\\
MPGMN & S0/S1 & 20.7 (24/18/20) & 23.0 (22/23/24) & 20.0 (24/18/18) & 16.7 (18/17/15)\\
Grid64 & S0 & 27.3 (29/27/26) & 27.7 (29/27/27) & 28.3 (31/28/26) & 28.0 (30/28/26)\\
Grid64 & S0/S1 & 26.3 (27/27/25) & 26.7 (27/28/25) & 27.7 (30/28/25) & 27.0 (27/30/24)\\
Grid256 & S0 & 26.7 (27/26/27) & 26.3 (26/26/27) & 26.0 (26/25/27) & 26.3 (26/26/27)\\
Grid256 & S0/S1 & 28.0 (27/30/27) & 27.3 (26/30/26) & 27.3 (26/30/26) & 27.0 (25/30/26)\\
WeightStats & S0 & 17.0 (20/14/17) & 13.0 (16/14/9) & 14.0 (18/9/15) & 13.3 (10/14/16)\\
WeightStats & S0/S1 & 17.0 (18/15/18) & 14.3 (17/12/14) & 15.7 (21/11/15) & 13.7 (13/14/14)\\
Dense CNN & Original RGB & 31.3 (31/31/32) & 32.7 (31/34/33) & 31.7 (30/34/31) & 31.7 (31/32/32)\\
Dense CNN & Original RGB repeat & 28.7 (32/32/22) & 27.7 (31/30/22) & 27.0 (29/31/21) & 28.0 (32/31/21)\\
\bottomrule\end{tabular}
\end{table*}

\begin{table*}[t]
\centering\small
\caption{Paired changes (percentage points, 95\% image-bootstrap intervals). Single-source rows compare fixed S0-trained readers across sources. Balanced gain compares S0/S1-trained with S0-trained readers on S2 at equal presentations and updates. Intervals condition on the fitted source bundle and trained seeds; these secondary comparisons are descriptive, without multiplicity-adjusted discovery claims.}
\label{tab:crossfit-changes}
\begin{tabular}{lrrr}\toprule
Reader & S2$-$S0 & I0$-$S0 & Balanced gain on S2\\\midrule
HiddenProbe & -1.7 [-3.7,+0.3] & -0.7 [-3.7,+2.3] & -1.7 [-6.3,+3.0]\\
ProbeGen & +0.7 [-0.3,+2.0] & +0.3 [-1.7,+2.3] & -3.3 [-7.3,+0.7]\\
MPGMN & -8.3 [-15.0,-1.7] & -6.7 [-14.3,+0.7] & +3.7 [-2.3,+9.7]\\
Grid64 & +1.0 [-0.3,+2.3] & +0.7 [-0.7,+2.0] & -0.7 [-2.7,+1.0]\\
Grid256 & -0.7 [-1.7,+0.0] & -0.3 [-1.0,+0.0] & +1.3 [-1.7,+4.3]\\
WeightStats & -3.0 [-8.7,+2.7] & -3.7 [-10.0,+2.7] & +1.7 [-2.3,+5.7]\\
Dense CNN & +0.3 [-1.0,+1.7] & +0.3 [-1.0,+1.7] & --\\
\bottomrule\end{tabular}
\end{table*}

\begin{table}[t]
\centering\small
\caption{Pilot reconstruction quality for all 600 images per source. No fits were dropped. PSNR uses RGB intensity range one, without clipping.}
\label{tab:crossfit-quality}
\begin{tabular}{lrrrr}\toprule
Source & Median & 10th pct. & Minimum & Below 30\\\midrule
S0 & 31.92 & 29.79 & 27.25 & 69\\
S1 & 31.37 & 29.26 & 26.79 & 118\\
S2 & 31.39 & 29.22 & 26.15 & 125\\
I0 & 31.70 & 29.65 & 26.89 & 93\\
\bottomrule\end{tabular}
\end{table}

\subsection{Primary comparison and interpretation}
The pre-specified contrast is the source-induced change in the
HiddenProbe--ProbeGen accuracy gap:
\begin{align*}
 &(\acc_{\mathrm{HP},\mathrm{S2}}-\acc_{\mathrm{PG},\mathrm{S2}})\\
 &\qquad -(\acc_{\mathrm{HP},\mathrm{S0}}-\acc_{\mathrm{PG},\mathrm{S0}}).
\end{align*}
Its estimate is $-2.33$ points with 95\% interval $[-4.67,0.00]$.
The pre-defined quality intersection with PSNR at least 30\,dB contains
78 images and gives $-3.12$ points $[-6.13,-0.25]$; it does not replace
the full-population primary result. MPGMN's S0-to-S2 drop is 8.3 points,
while the designated original-RGB CNN comparison changes by 0.3 points.

The prospective expansion rule required an absolute gap change of at
least five points, or a ten-point reader drop on S2/I0 accompanied by
a dense-image change no greater than one third as large. Neither
criterion was met. This was an experimental allocation rule, not a
significance threshold or evidence of general source robustness.
The later orbit study fixes a different, exact-function question and
does not revise this pilot decision. Complete results include the
larger S1 graph-reader drop and both CNN repetitions.

\section{Language-Model Readers and Interface Controls}\label{app:lm}
\subsection{Shared input and scoring}
The weight prompt specifies the two-dimensional coordinates, shallow
ReLU architecture, nonpositive foreground convention, six shape
definitions, and all weights as six-significant-digit JSON. Each origin
has a randomized A--F menu shared across conditions. Readers receive
no demonstrations or target-function executions. A redacted condition
removes only the weights; an explicit-cue condition replaces them with
the target shape name. The cue tests instruction and answer mapping,
while also shortening the input, so it does not isolate the mechanism
of weight-reading failure.

\subsection{Deterministic next-token readers}
The public checkpoints are Qwen2.5-0.5B/1.5B-Instruct
\cite{qwen2025qwen25}, SmolLM2-360M-Instruct \cite{allal2025smol}, and
Qwen3.5-0.8B \cite{qwen2026qwen35}. They use supplied chat templates
and FP32 inference through Transformers \cite{wolf2020transformers};
Qwen3.5 uses its text-only non-thinking template. Each candidate letter
is a verified single token. The answer is the maximum next-token
candidate score, without sampling or fine-tuning.

Contextual calibration subtracts the corresponding menu-matched redacted
log probability with fixed coefficient one \cite{zhao2021calibrate}.
Candidate entropy is computed after renormalization over A--F, and
is not full-vocabulary entropy. The unrestricted best token is retained
as a diagnostic. Identical canonical strings share one deterministic
score; they are not counted as independent repeated calls.

\begin{table*}[t]
\centering\small
\caption{Complete original and contextual-calibration results on the same 60 origins. Intervals are Wilson intervals; disagreement and both-correct are percentages.}\label{tab:lmfull}
\begin{tabular}{llrrrrr}
\toprule
Model & Rule & Raw [95\% CI] & Joint [95\% CI] & Canon. [95\% CI] & Dis. & Both \\
\midrule
Qwen2.5 0.5B & original & 21.7 [13.1, 33.6] & 21.7 [13.1, 33.6] & 21.7 [13.1, 33.6] & 0.0 & 21.7 \\
Qwen2.5 0.5B & calibrated & 21.7 [13.1, 33.6] & 21.7 [13.1, 33.6] & 21.7 [13.1, 33.6] & 0.0 & 21.7 \\
Qwen2.5 1.5B & original & 18.3 [10.6, 29.9] & 16.7 [9.3, 28.0] & 16.7 [9.3, 28.0] & 6.7 & 16.7 \\
Qwen2.5 1.5B & calibrated & 21.7 [13.1, 33.6] & 23.3 [14.4, 35.4] & 23.3 [14.4, 35.4] & 3.3 & 21.7 \\
SmolLM2 0.36B & original & 21.7 [13.1, 33.6] & 21.7 [13.1, 33.6] & 21.7 [13.1, 33.6] & 1.7 & 21.7 \\
SmolLM2 0.36B & calibrated & 23.3 [14.4, 35.4] & 23.3 [14.4, 35.4] & 23.3 [14.4, 35.4] & 5.0 & 23.3 \\
Qwen3.5 0.8B & original & 15.0 [8.1, 26.1] & 16.7 [9.3, 28.0] & 13.3 [6.9, 24.2] & 13.3 & 11.7 \\
Qwen3.5 0.8B & calibrated & 13.3 [6.9, 24.2] & 13.3 [6.9, 24.2] & 13.3 [6.9, 24.2] & 0.0 & 13.3 \\
\bottomrule
\end{tabular}
\end{table*}

\begin{table}[t]
\centering\small
\caption{\textbf{Serializer components, Qwen2.5 0.5B.} Percent accuracy and the number of predictions changing from the original among 60 origins. Component ablations process the same joint weights. Zero changes are an observed sample result, not a population-risk certificate. The main paper reports its raw-accuracy interval.}\label{tab:ablation}
\begin{tabular}{lrr}
\toprule
Representation & Acc. & Changed /60 \\
\midrule
Original & 21.7 & 0 \\
Permutation & 21.7 & 0 \\
Scaling & 21.7 & 0 \\
Joint & 21.7 & 0 \\
Joint then normalize & 21.7 & 0 \\
Joint then sort & 21.7 & 0 \\
Canonical & 21.7 & 0 \\
\bottomrule
\end{tabular}
\end{table}

These controls include \NumLMQueries\ scored prompts. Qwen2.5-0.5B
chooses F for all original inputs, matching its redacted choices.
Qwen2.5-1.5B recognizes 58/60 explicit cues but only 11/60 original
weight inputs. Stable output and correct answer mapping therefore
provide different evidence from successful interpretation of weights.

\section{Generated-Answer Experiments}\label{app:frontier}
\subsection{Configurations and inclusion}
The exploratory configurations are \texttt{gpt-6-astra},
\texttt{gpt-6.1-sol}, \texttt{gpt-6.1-terra}, and \texttt{gpt-6-luna},
requested with high reasoning effort. These identifiers describe the
requested routes, not verified backend snapshots. No temperature,
sampling seed, or equalized reasoning-token budget is assumed.
Each response uses a fresh session without tools or demonstrations.
Model context excludes gold labels and experiment identifiers.

Every configuration receives 60 origins under eight conditions:
original, original repeat, permutation, joint, canonical, canonical
repeat, redacted, and cue. Repeated prompts are textually identical;
condition order is shuffled. A valid answer contains exactly one
uppercase A--F after trimming whitespace. Invalid completions count as
errors. The earliest retained completed answer for each scheduled input
is selected independently of its validity or correctness.

The analyzed collection contains 1,920 distinct sessions. Later
completed duplicates are excluded. Invocations without retained responses
cannot be ordered against retained completions and contribute no scores;
the collection ledger records these exclusions. All weight-bearing and
repeat responses are valid. Redacted-condition invalid counts are
1, 3, 54, and 0 for Astra, Sol, Terra, and Luna.

\begin{table*}[t]
\centering\small
\caption{\textbf{Every generated-answer condition.} All completed invalid outputs are included in the denominator. The two repeats are fresh calls, not copied answers. No condition is selected by its score.}\label{tab:frontier_full}
\begin{tabular}{lrrrr}
\toprule
Requested model & Condition & Correct /60 & Accuracy [95\% CI] & Invalid /60 \\
\midrule
Astra 6 & raw & 52/60 & 86.7 [75.8, 93.1] & 0 \\
Astra 6 & raw\_repeat & 54/60 & 90.0 [79.9, 95.3] & 0 \\
Astra 6 & perm & 51/60 & 85.0 [73.9, 91.9] & 0 \\
Astra 6 & joint & 47/60 & 78.3 [66.4, 86.9] & 0 \\
Astra 6 & canonical & 51/60 & 85.0 [73.9, 91.9] & 0 \\
Astra 6 & canonical\_repeat & 55/60 & 91.7 [81.9, 96.4] & 0 \\
Astra 6 & redacted & 10/60 & 16.7 [9.3, 28.0] & 1 \\
Astra 6 & textcue & 60/60 & 100.0 [94.0, 100.0] & 0 \\
Sol 6.1 & raw & 50/60 & 83.3 [72.0, 90.7] & 0 \\
Sol 6.1 & raw\_repeat & 52/60 & 86.7 [75.8, 93.1] & 0 \\
Sol 6.1 & perm & 54/60 & 90.0 [79.9, 95.3] & 0 \\
Sol 6.1 & joint & 49/60 & 81.7 [70.1, 89.4] & 0 \\
Sol 6.1 & canonical & 54/60 & 90.0 [79.9, 95.3] & 0 \\
Sol 6.1 & canonical\_repeat & 51/60 & 85.0 [73.9, 91.9] & 0 \\
Sol 6.1 & redacted & 9/60 & 15.0 [8.1, 26.1] & 3 \\
Sol 6.1 & textcue & 60/60 & 100.0 [94.0, 100.0] & 0 \\
Terra 6.1 & raw & 42/60 & 70.0 [57.5, 80.1] & 0 \\
Terra 6.1 & raw\_repeat & 36/60 & 60.0 [47.4, 71.4] & 0 \\
Terra 6.1 & perm & 46/60 & 76.7 [64.6, 85.6] & 0 \\
Terra 6.1 & joint & 29/60 & 48.3 [36.2, 60.7] & 0 \\
Terra 6.1 & canonical & 42/60 & 70.0 [57.5, 80.1] & 0 \\
Terra 6.1 & canonical\_repeat & 43/60 & 71.7 [59.2, 81.5] & 0 \\
Terra 6.1 & redacted & 1/60 & 1.7 [0.3, 8.9] & 54 \\
Terra 6.1 & textcue & 60/60 & 100.0 [94.0, 100.0] & 0 \\
Luna 6 & raw & 21/60 & 35.0 [24.2, 47.6] & 0 \\
Luna 6 & raw\_repeat & 18/60 & 30.0 [19.9, 42.5] & 0 \\
Luna 6 & perm & 18/60 & 30.0 [19.9, 42.5] & 0 \\
Luna 6 & joint & 19/60 & 31.7 [21.3, 44.2] & 0 \\
Luna 6 & canonical & 20/60 & 33.3 [22.7, 45.9] & 0 \\
Luna 6 & canonical\_repeat & 18/60 & 30.0 [19.9, 42.5] & 0 \\
Luna 6 & redacted & 9/60 & 15.0 [8.1, 26.1] & 0 \\
Luna 6 & textcue & 60/60 & 100.0 [94.0, 100.0] & 0 \\
\bottomrule
\end{tabular}
\end{table*}

\begin{table*}[t]
\centering\small
\caption{\textbf{Paired generated-answer diagnostics.} Counts out of 60 origins; $\Delta$ is second minus first accuracy in percentage points. Intervals are class-stratified origin-bootstrap percentiles. Any invalid member is a disagreement, even when both members are invalid. Raw/canonical and prior/raw contrasts are descriptive follow-up summaries; the full output is retained.}\label{tab:frontier_pairs}
\begin{tabular}{lrrrrrr}
\toprule
Requested model & First/second & Dis. & Harm & Help & Wrong flip & $\Delta$ acc. [95\% CI] \\
\midrule
Astra 6 & Raw/Perm. & 9 & 4 & 3 & 2 & -1.7 [-10.0, +6.7] \\
Astra 6 & Raw/Joint & 12 & 8 & 3 & 1 & -8.3 [-18.3, +1.7] \\
Astra 6 & Raw/Repeat & 5 & 1 & 3 & 1 & +3.3 [-1.7, +10.0] \\
Astra 6 & Canon./Repeat & 7 & 1 & 5 & 1 & +6.7 [+0.0, +13.3] \\
Astra 6 & Raw/Canon. & 9 & 5 & 4 & 0 & -1.7 [-11.7, +8.3] \\
Astra 6 & Prior/Raw & 47 & 0 & 42 & 5 & +70.0 [+61.7, +78.3] \\
Sol 6.1 & Raw/Perm. & 8 & 1 & 5 & 2 & +6.7 [+0.0, +13.3] \\
Sol 6.1 & Raw/Joint & 11 & 5 & 4 & 2 & -1.7 [-11.7, +8.3] \\
Sol 6.1 & Raw/Repeat & 9 & 3 & 5 & 1 & +3.3 [-5.0, +13.3] \\
Sol 6.1 & Canon./Repeat & 7 & 5 & 2 & 0 & -5.0 [-13.3, +3.3] \\
Sol 6.1 & Raw/Canon. & 13 & 4 & 8 & 1 & +6.7 [-3.3, +18.3] \\
Sol 6.1 & Prior/Raw & 48 & 0 & 41 & 7 & +68.3 [+58.3, +76.7] \\
Terra 6.1 & Raw/Perm. & 11 & 2 & 6 & 3 & +6.7 [-1.7, +15.0] \\
Terra 6.1 & Raw/Joint & 21 & 16 & 3 & 2 & -21.7 [-35.0, -10.0] \\
Terra 6.1 & Raw/Repeat & 16 & 10 & 4 & 2 & -10.0 [-21.7, +1.7] \\
Terra 6.1 & Canon./Repeat & 17 & 7 & 8 & 2 & +1.7 [-10.0, +13.3] \\
Terra 6.1 & Raw/Canon. & 13 & 5 & 5 & 3 & +0.0 [-8.3, +8.3] \\
Terra 6.1 & Prior/Raw & 58 & 0 & 41 & 17 & +68.3 [+60.0, +78.3] \\
Luna 6 & Raw/Perm. & 24 & 9 & 6 & 9 & -5.0 [-16.7, +8.3] \\
Luna 6 & Raw/Joint & 25 & 10 & 8 & 7 & -3.3 [-16.7, +10.0] \\
Luna 6 & Raw/Repeat & 23 & 7 & 4 & 12 & -5.0 [-15.0, +5.0] \\
Luna 6 & Canon./Repeat & 19 & 6 & 4 & 9 & -3.3 [-13.3, +6.7] \\
Luna 6 & Raw/Canon. & 21 & 5 & 4 & 12 & -1.7 [-11.7, +8.3] \\
Luna 6 & Prior/Raw & 43 & 3 & 15 & 25 & +20.0 [+10.0, +30.0] \\
\bottomrule
\end{tabular}
\end{table*}

\begin{table}[!htbp]
\centering\small
\caption{\textbf{Joint disagreement above the identical-raw baseline.} Percentage-point differences, computed within each origin before class-stratified bootstrap resampling. Negative values are possible. An interval containing zero does not establish equivalence or absence of representation sensitivity.}\label{tab:frontier_excess}
\begin{tabular}{lrr}
\toprule
Requested model & $D_j-D_r$ & 95\% CI \\
\midrule
Astra 6 & +11.7 & [+1.7, +20.0] \\
Sol 6.1 & +3.3 & [-5.0, +11.7] \\
Terra 6.1 & +8.3 & [-5.0, +23.3] \\
Luna 6 & +3.3 & [-10.0, +18.3] \\
\bottomrule
\end{tabular}
\end{table}

\subsection{Recognition, repeats, and exploratory inference}
Original-weight accuracy ranges from 35.0\% to 86.7\%; every explicit
cue is correct. Joint accuracy is 78.3\% for Astra, 81.7\% for Sol,
48.3\% for Terra, and 31.7\% for Luna. Raw/joint disagreement ranges
from 18.3\% to 41.7\%, compared with 8.3--38.3\% between original
repeats. Canonical repeats disagree on 11.7--31.7\% despite identical
input strings.

\begin{figure}[t]
\centering\includegraphics[width=\linewidth]{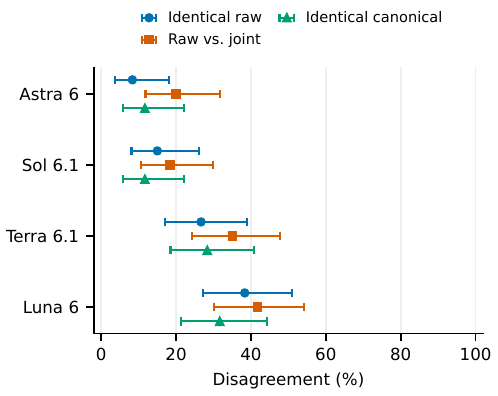}
\caption{\textbf{Repeated-input variation supplies the stochastic reference.}
Disagreement rates on the same 60 origins, with Wilson intervals.
Each repeat is a fresh answer to an identical prompt. Differences
between rates are evaluated as paired origin-level contrasts.}
\label{fig:frontier_repeats}
\end{figure}

For Astra, raw/joint minus raw-repeat disagreement is $+11.7$ points,
with descriptive bootstrap interval $[+1.7,+20.0]$. The exact paired
sensitivity test instead gives $p=0.0654$, or 0.2617 after adjustment
within the four-configuration family. Terra loses 16 correct answers
and gains three on the raw/joint pair ($p=0.00443$), but using its raw
repeat as the reference gives $p=0.189$. These retrospective checks
show why the observed repeat baseline and test definition must be
reported rather than used to rank general model sensitivity.

\begin{table*}[t]
\centering\small
\caption{\textbf{Exact paired sensitivity checks.} Two-sided, unadjusted values from the retained 60 origins. The first three columns compare accuracy; the last compares paired disagreement indicators. All comparisons are retrospective and exploratory.}\label{tab:exact}
\begin{tabular}{lrrrr}\toprule
Configuration & Raw/Joint & Raw repeat/Joint & Raw/Canonical & Joint excess \\
\midrule
Astra 6 & 0.2266 & 0.0654 & 1.0000 & 0.0654 \\
Sol 6.1 & 1.0000 & 0.5078 & 0.3877 & 0.7266 \\
Terra 6.1 & 0.0044 & 0.1892 & 1.0000 & 0.4049 \\
Luna 6 & 0.8145 & 1.0000 & 1.0000 & 0.8318 \\
\bottomrule\end{tabular}
\end{table*}

For binary paired indicators, the two-sided conditional sensitivity
value with discordance counts $b,c$ is
\[
 \min\left\{1,2\sum_{j=0}^{\min(b,c)}{b+c\choose j}2^{-(b+c)}\right\}.
\]
It assumes independent equiprobable discordance directions under its
null. Bootstrap intervals instead resample origins, carrying all
answers together. Neither inference rule establishes distributional
invariance from a small repeat-subtracted difference.

\subsection{Prospective two-configuration collection}
Astra and Sol were selected for their stronger original recognition
before the new collection. Each receives three fresh raw and three
fresh joint responses per origin, giving 720 responses in total.
Prompts, menus, numerical inputs, inclusion rules, and the two-test
Holm family are fixed before collection. Models and conditions are
interleaved in randomized order. No historical response contributes
to this prospective estimate.

The primary tests compare full response laws with the exact categorical
procedure derived below. Accuracy changes and origin-bootstrap
intervals are secondary. Invalid answers would enter the distributional
statistic as a seventh category and accuracy as errors. Independent,
stationary calls are assumptions of the test; fresh sessions implement
the design without certifying server stationarity or excluding shared
temporal effects.

\section{A Known-Law Next-Token Control}\label{app:open_large}
Qwen3.5-27B \cite{qwen2026large} supplies a separate 300-evaluation
control: the same 60 origins under raw, joint, canonical, redacted,
and cue inputs. Text-only inference uses the non-thinking template,
BF16 weights, and FP32 vocabulary log-softmax. All six letters are
single tokens. Candidate selection measures restricted next-token
recognition, not free-form generated-answer accuracy.

Renormalizing the six scores gives known categorical laws at each
origin. We use them to calculate raw/joint discrepancy and simulate
1,000 two-repeat and 1,000 three-repeat experiments. The null samples
the raw law on both sides; the alternative samples raw and joint laws.
Every simulated test uses the same within-origin exact allocation
procedure. This gives a power measurement for the observed candidate
laws, not a power estimate for the API readers.

\begin{table}[t]
\centering\footnotesize
\caption{The 27B candidate-score control uses the fixed 60-origin menu. Mass is the median full-vocabulary probability assigned to A--F; calibration subtracts menu-matched redacted scores.}\label{tab:qwen27}
\begin{tabular}{lrrr}
\toprule
Input & Accuracy (\%) & Calibrated (\%) & Mass (\%)\\
\midrule
raw & 11.7 & 21.7 & 99.93\\
joint & 15.0 & 18.3 & 99.93\\
canonical & 15.0 & 26.7 & 99.93\\
redacted & 15.0 & -- & 99.93\\
textcue & 100.0 & -- & 99.99\\
\bottomrule
\end{tabular}
\end{table}

\begin{table}[t]
\centering\footnotesize
\caption{Known candidate-law simulation: rejection rate (\%, 95\% binomial interval) over 1,000 trials per design and law. Nominal level is 5\%; this is a separate, single-test simulation.}\label{tab:power27}
\begin{tabular}{lcc}
\toprule
Repeats per side & Null rejection & Alternative rejection\\
\midrule
2 & 3.8 [2.8, 5.2] & 7.0 [5.6, 8.8]\\
3 & 3.7 [2.7, 5.1] & 11.5 [9.7, 13.6]\\
\bottomrule
\end{tabular}
\end{table}

Raw, joint, and canonical accuracies are 7/60, 9/60, and 9/60, versus
9/60 redacted and 60/60 with explicit cues. Candidate letters carry
at least 0.9983 of the next-token mass in every condition. The mean
half-squared discrepancy is 0.01350 and mean total variation is 0.13901.
Alternative rejection is 7.0\% with two repeats and 11.5\% with three;
null rejection is 3.8\% and 3.7\%, respectively. Binomial intervals
quantify simulation error. The result demonstrates limited power at
these laws and replication levels, without identifying a universal
sample-size requirement or a causal effect of model size.

\section{Replication and Exact Conditional Testing}\label{app:replication}
\subsection{The comparison preserves origin identity}
For $n$ fixed origins, let $p_i,q_i\in\R^K$ be the class-response
probabilities for two representations. The estimand is
\[
 S_n=\frac{1}{2n}\sum_i\|p_i-q_i\|_2^2.
\]
It vanishes exactly when the laws agree at every selected origin.
Pooling labels before comparison targets a different quantity:
\[
 \tfrac12\|n^{-1}\textstyle\sum_i(p_i-q_i)\|_2^2\leq S_n.
\]
For example, $p_1=q_2=(1,0)$ and $p_2=q_1=(0,1)$ have identical
pooled laws but $S_2=1$. The audit must therefore retain origin identity.
The categorical target also requires literal equality of categories:
if invalid answers are represented by an additional symbol, two such
symbols collide. The behavioral convention that every invalid pair
disagrees is not used for this distributional statistic.

\subsection{Why one response per transformed input is insufficient}
The following specialization of the random-singleton construction in
multi-distribution testing \cite{aliakbarpour2020multiple} states a
worst-case limit for unrestricted origin-specific laws.

\begin{proposition}[One transformed response per origin]
\label{prop:one_repeat}
Suppose $K\geq2$ and each fixed origin supplies one transformed response.
Any level-$\alpha$ test of $p_i=q_i$ for every $i$ has power at most
$\alpha$ at some alternative with $S_n=(1-1/K)/2$, even if each $p_i$
is known and arbitrarily many original-input repeats are available.
\end{proposition}
\begin{proof}
Set all $p_i$ to the uniform law $u_K$. Under the null set $q_i=u_K$.
For the alternative, independently draw a uniform label $J_i$ and fix
$q_i=\delta_{J_i}$. Every fixed alternative has
$\|u_K-\delta_{J_i}\|^2/2=(1-1/K)/2$. Averaging over the $J_i$ makes
the single transformed responses independent uniform labels, exactly
as under the null. Original-input responses also have the same law
in both cases. The entire observation law is thus identical for the
null and the alternative mixture. Average rejection is at most
$\alpha$, so at least one fixed alternative has power at most $\alpha$.
\end{proof}

This rules out a uniform power guarantee for this design, not detection
of every alternative. Assumptions linking origins or other allocations
of repeats define different problems. There is also an unbiasedness
obstruction: with one response from an arbitrary $q_i$, the expectation
of an integrable statistic is affine in $q_i$, whereas $S_n$ contains
$\|q_i\|^2/2$. The same obstruction applies with one raw response.
Accuracy change is linear in the response laws and remains estimable.

RUT uses reference-model token scores to detect API substitutions with
one target response per prompt \cite{zhu2026rut}. Such power against
particular alternatives is compatible with \cref{prop:one_repeat}, which
rules out a uniform guarantee over unrestricted origin-specific laws.
The proposition already allows known reference laws; additional reference
access alone does not remove its least-favorable alternatives.

\subsection{Two-sided repeats estimate the discrepancy}
With independent $X_{i1},X_{i2}\sim p_i$ and $Z_{i1},Z_{i2}\sim q_i$,
write
\begin{align*}
 T_i={}&\tfrac12\left(\ind\{X_{i1}=X_{i2}\}+\ind\{Z_{i1}=Z_{i2}\}\right)\\
 &-\tfrac14\sum_{a,b=1}^2\ind\{X_{ia}=Z_{ib}\}.
\end{align*}
The collision expectations give
$\E T_i=(\|p_i\|^2+\|q_i\|^2)/2-p_i^\top q_i$.
Consequently $n^{-1}\sum_iT_i$ is unbiased for $S_n$.
This is half the unbiased delta-kernel MMD-squared estimator,
matched by origin \cite{gretton2012kernel,gao2025equality}.
A finite-sample estimate can be negative and is not clipped.
Since $T_i\in[-1/2,1]$, independent origin clusters satisfy
\[
 \Pr(|\widehat S_n-S_n|\geq\epsilon)
 \leq2\exp(-8n\epsilon^2/9).
\]
The bound conditions on the fixed origins; they need not share a law.
It is distinct from uncertainty obtained by resampling origins.

\subsection{Exact two-repeat null from collision patterns}
Under equality and independent stationary calls, the four responses
within each origin are exchangeable. Conditional on their multiset,
the six assignments of two positions to each condition are equally
likely. Complementary assignments give the same $T_i$. The possible
multiplicity patterns therefore determine its full conditional law:
\begin{center}
\small
\begin{tabular}{lcc}
\toprule
Pooled multiplicities & Statistic & Probability\\
\midrule
$(4),(3,1),(1,1,1,1)$ & $0$ & $1$\\
$(2,2)$ & $1,\;-1/2$ & $1/3,\;2/3$\\
$(2,1,1)$ & $1/2,\;-1/4$ & $1/3,\;2/3$\\
\bottomrule
\end{tabular}
\end{center}
If $A,B$ count origins of the last two types, then
\begin{gather*}
 4n\widehat S_n^*=3(2U+V)-(2A+B),\\
 U\sim\operatorname{Bin}(A,1/3),\qquad
 V\sim\operatorname{Bin}(B,1/3),
\end{gather*}
independently. For observed $s=4n\widehat S_n$, the exact upper tail is
\[
 \frac{\displaystyle
 \sum_{\substack{0\leq u\leq A,\,0\leq v\leq B\\3(2u+v)-(2A+B)\geq s}}
 {A\choose u}{B\choose v}2^{A+B-u-v}}
 {3^{A+B}}.
\]
Its conditional variance is $(4A+B)/(8n^2)$ and its smallest attainable
inclusive tail is $3^{-(A+B)}$. Constant patterns supply conditioning
information but cannot change the permutation statistic.
The product randomization requires independence of origin clusters,
or the corresponding joint invariance under within-origin assignments.
No responses or menus are exchanged between origins.

The exact test is conservative when its discrete support cannot attain
the nominal level. A bounded-sum argument gives
$p_{\rm cond}\leq\exp(-8nt^2/9)$ for observed $t>0$.
For $c_{n,\alpha}=\sqrt{9\log(1/\alpha)/(8n)}$ and alternatives
$S_n\geq\delta>c_{n,\alpha}$, rejection probability is at least
$1-\exp[-8n(\delta-c_{n,\alpha})^2/9]$.
This establishes consistency at fixed positive separation as the
number of origins grows; it is not a power estimate for the present
60-origin experiment.

\subsection{Exploratory original/canonical comparisons}
The historical collection provides two responses per side for original
and canonical inputs. It does not provide a second joint response,
so this analysis cannot estimate the joint-input discrepancy.
All four configurations and all 60 origins are retained.

\begin{table*}[t]
\centering\small\setlength{\tabcolsep}{6pt}
\caption{\textbf{Recognition loss and distributional evidence in 720 fresh responses.}
Three raw and three joint responses for each of 60 origins per configuration.
Accuracy changes have descriptive 95\% origin-bootstrap intervals.
$\widehat S_n$ estimates the mean half-squared response-law distance;
$p_H$ is the exact distribution-test value after the pre-specified two-test
Holm adjustment.}\label{tab:fresh}
\begin{tabular*}{\textwidth}{@{\extracolsep{\fill}}lrrrrr@{}}
\toprule
Reader & Raw (\%) & Joint (\%) & Change (pp), 95\% interval & $\widehat S_n$ & $p_H$\\
\midrule
Astra & 88.3 & 81.1 & $-7.2\;[-12.8,-2.2]$ & 0.0185 & 0.162\\
Sol & 86.7 & 79.4 & $-7.2\;[-12.8,-1.7]$ & 0.0259 & 0.162\\
\bottomrule
\end{tabular*}
\end{table*}

\begin{table}[t]
\centering\small\setlength{\tabcolsep}{3pt}
\caption{\textbf{Raw/canonical distribution comparison.} Retrospective
analysis on 60 origins. $I$ counts origins with nonconstant conditional
permutation terms. $p$ is exact and one-sided; H4/H16 denote Holm
adjustment in the four-test and 16-test families. Negative estimates
are retained.}\label{tab:distribution}
\begin{tabular}{lrrrrr}
\toprule
Config. & $\widehat S_n$ & $I$ & $p$ & H4 & H16\\
\midrule
Astra & 0.0208 & 5 & 0.2263 & 0.6024 & 1.0000 \\
Sol & 0.0750 & 7 & 0.0087 & 0.0348 & 0.1303 \\
Terra & -0.0417 & 12 & 0.9615 & 0.9615 & 1.0000 \\
Luna & 0.0375 & 16 & 0.2008 & 0.6024 & 1.0000 \\
\bottomrule
\end{tabular}
\end{table}

\begin{table}[t]
\centering\small\setlength{\tabcolsep}{3pt}
\caption{\textbf{Conditional information in retained repeats.} Counts
of pooled four-response multiplicities. Only $(2,2)$ and $(2,1,1)$
have nonconstant permutation terms; $p_{\min}=3^{-(A+B)}$.}
\label{tab:collision}
\begin{tabular}{lrrrrrr}
\toprule
Config. & 4 & 3+1 & 2+2 & 2+1+1 & 1+1+1+1 & $p_{\min}$\\
\midrule
Astra & 49 & 6 & 2 & 3 & 0 & $3^{-5}$ \\
Sol & 41 & 12 & 5 & 2 & 0 & $3^{-7}$ \\
Terra & 37 & 11 & 4 & 8 & 0 & $3^{-12}$ \\
Luna & 25 & 17 & 8 & 8 & 2 & $3^{-16}$ \\
\bottomrule
\end{tabular}
\end{table}

Astra, Sol, Terra, and Luna have 5, 7, 12, and 16 origins with
nonconstant terms. Sol has $\widehat S_n=0.0750$ and exact $p=0.00869$,
while its first original/canonical accuracy change has an interval
spanning zero. The two comparisons target different quantities.
Holm adjustment gives 0.0348 across the four distribution tests and
0.1303 across a broader sixteen-test family comprising original/joint
accuracy, original/canonical accuracy, excess disagreement, and
original/canonical distribution for each configuration. Both analyses
are retrospective. They do not establish a confirmatory recognition
benefit from canonicalization.

\subsection{Three responses per side}
For $m\geq2$, the same argument gives
\begin{align*}
 T_i^{(m)}={}&\frac{1}{2\binom m2}\sum_{a<b}
 (\ind\{X_{ia}=X_{ib}\}+\ind\{Z_{ia}=Z_{ib}\})\\
 &-\frac{1}{m^2}\sum_{a,b}\ind\{X_{ia}=Z_{ib}\},
\end{align*}
with expectation $\|p_i-q_i\|^2/2$.
For three responses per side, $18T_i=3W_i-2C_i$, where $W_i$ counts
six within-condition collisions and $C_i$ counts nine cross-condition
collisions. Enumerating the twenty balanced allocations per origin
and convolving their integer laws gives the exact test, with denominator
$20^{60}$ before reduction. Reused pairs share responses and are not
independent samples.

For a pooled $(5,1)$ pattern, every allocation gives
$18T_i=3(4)-2(6)=0$. In the prospective experiment, Astra has forty
$(6)$ and eleven $(5,1)$ origins, leaving nine informative terms.
Sol has thirty-four and thirteen, leaving thirteen. A gold label can
make the direction of an accuracy change visible even when this
omnibus statistic is constant. This explains how the secondary accuracy
intervals can exclude zero while neither primary distribution test
rejects. The test remains an origin-matched conditional MMD procedure,
with the same independence and stationarity assumptions as its derivation.

\section{Relation to Post-hoc Symmetrization and Symmetry Priors}
Post-hoc symmetrization and our finite transformation audit address
related but distinct questions. Zhou \emph{et al.} \cite{zhou2026posthoc}
explain why normalized Haar averaging is unavailable for non-compact
groups and study Weyl, orthogonal-projection, and frame alternatives.
Their orthogonal projection can improve prediction; benefits of the
other constructions depend on the setting. Integer phase shifts and
positive ReLU rescalings give non-compact components in the parameter
actions considered here. Our finite transformation averages are not
normalized Haar projections, so compact-group guarantees do not apply
automatically. Non-compactness alone does not imply an accuracy loss.

Adly \cite{adly2026exchange} reports exact agreement between test-time
orbit averaging and an equivariant model in a particular finite-group
experiment. This is a result about predictor outputs, not a requirement
that a reader consume neural-field outputs. A hidden-feature reader also
produces a probability vector that can be pooled, as in our experiments.
The different group, sampled action design, training distribution, and
architectures prevent transferring that empirical equivalence directly.

Our results show both signs of averaging effects: CIFAR10 post-sine
features gain 1.6 points in the pilot, while preactivation features lose
10.4 points. The orbit-variance shares quantify sensitivity under the
specified design. Without labels they do not determine whether changed
predictions become more or less correct, and we do not propose a
threshold for enabling averaging. Feature-location comparisons share
head depth and hidden width, but their unequal input dimensions and
information content prevent an equal-capacity causal interpretation.

\section{Reproducibility Scope}
The retained CPU package contains fitted fields, splits, reader outputs,
candidate scores, prompts, and generated-response records. The source
pilot includes its exported checkpoints. The larger SIREN study retains
raw fields and reader states remotely; the local package contains its
fixed protocols, image manifest, verified aggregate analyses, and
regeneration code. This distinction limits local replay of those
particular trained states. Tables and the new diagnostic figures can
be regenerated from the retained measurements without new inference.
Dependency versions and upstream revisions are documented with the code,
following the separation of experimental evidence and execution records
advocated by reproducibility practice \cite{pineau2021repro}.

\clearpage
\onecolumn
\section{Visual Reconstructions and Precision Diagnostics}
The visual examples are selected by reconstruction quality rather than
reader success. Each shape is the median-IoU test origin in its class;
each digit is the median-MSE test origin. The distributions include all
480 shape and 1,000 digit fields, including imperfect reconstructions.
\begin{center}
\includegraphics[width=.9\textwidth]{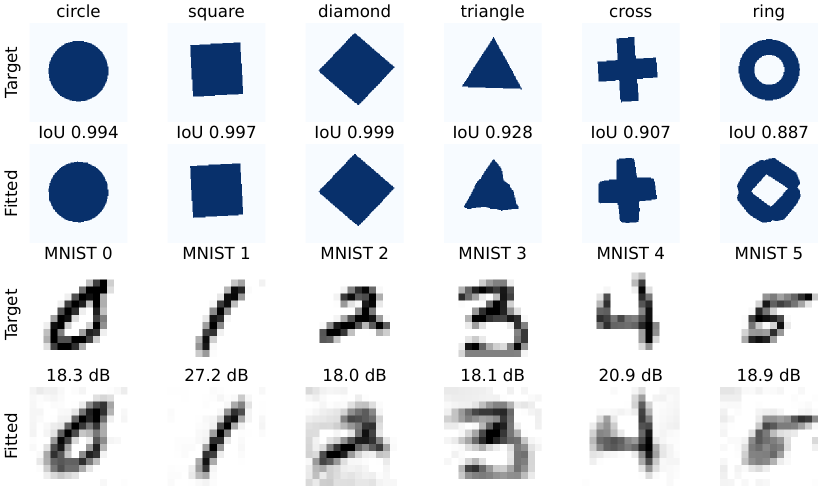}
\captionof{figure}{\textbf{Targets and the functions fitted to them.}
Representative shape fields and digit reconstructions. The same visual
origin supplies every transformed weight representation in an audit.}
\label{fig:examples}
\medskip
\includegraphics[width=\textwidth]{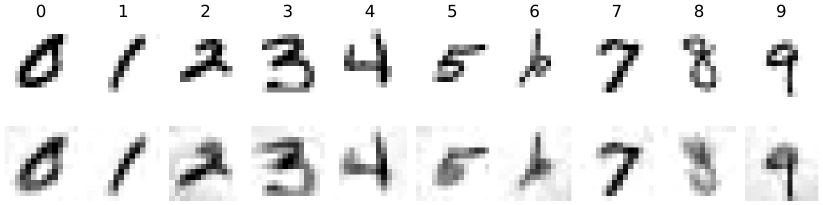}
\captionof{figure}{\textbf{Reconstructions across all ten digit classes.}
The source image is above its fitted field. Selection uses the
within-class median test reconstruction error.}
\end{center}
\clearpage
\begin{center}
\includegraphics[width=\textwidth]{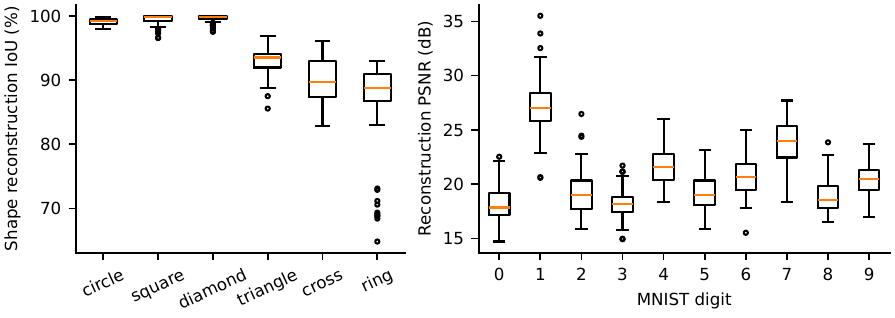}
\captionof{figure}{\textbf{Fitting quality over the complete populations.}
Boxes show quartiles and medians, with 1.5-IQR whiskers and retained
outliers. Reader evaluation excludes no low-quality field.}
\bigskip
\includegraphics[width=.65\textwidth]{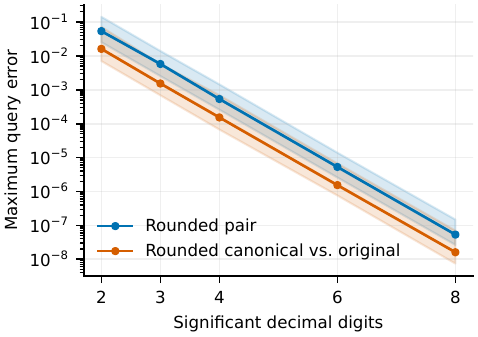}
\captionof{figure}{\textbf{Serialization error decreases with numerical precision.}
Continuously scaled shape fields with exponents in $[-4,4]$.
Curves show the median per-origin maximum query error; bands show
the 5th--95th percentiles. Rounded-pair error compares separately
printed equivalents, while canonical error compares the printed
canonical field with its full-precision origin.}
\label{fig:numeric}
\end{center}
\twocolumn

\section{Released-Configuration CIFAR10 Audit}
\label{app:official_cifar}
\begin{table*}[t]
\centering\small\setlength{\tabcolsep}{3pt}
\caption{\textbf{Official CIFAR10 audit by reader seed.} All rates are percentages on the same 10,000 test origins. Native uses released inference precision; Raw and transformed accuracy use FP64 target evaluation with FP32 readouts. $D_1,D_3,D_s$ are disagreements with the audit Raw prediction; $D_p$ compares native and FP32 clean inference.}\label{tab:official_seeds}
\begin{tabular*}{\textwidth}{@{\extracolsep{\fill}}llrrrrrrrrr@{}}\toprule
Reader & Seed & Native & Raw & Phase 1 & Phase 3 & Sign & $D_1$ & $D_3$ & $D_s$ & $D_p$\\
\midrule
HiddenProbe & 0 & 57.73 & 57.83 & 57.42 & 53.82 & 57.82 & 7.03 & 22.98 & 10.14 & 2.19 \\
HiddenProbe & 1 & 58.25 & 58.16 & 58.01 & 54.76 & 58.38 & 7.99 & 22.90 & 11.16 & 2.37 \\
HiddenProbe & 2 & 57.47 & 57.52 & 57.38 & 54.69 & 57.60 & 6.81 & 20.22 & 10.25 & 2.24 \\
ProbeGen & 0 & 57.36 & 57.36 & 57.36 & 57.36 & 57.36 & 0.00 & 0.00 & 0.00 & 0.00 \\
ProbeGen & 1 & 56.62 & 56.62 & 56.62 & 56.62 & 56.62 & 0.00 & 0.00 & 0.00 & 0.00 \\
ProbeGen & 2 & 57.86 & 57.86 & 57.86 & 57.86 & 57.86 & 0.00 & 0.00 & 0.00 & 0.00 \\
\bottomrule\end{tabular*}
\end{table*}
\begin{table*}[t]
\centering\small\setlength{\tabcolsep}{5pt}
\caption{\textbf{Paired accuracy changes in the official corpus.} Means over 3 reader seeds per method. Intervals are descriptive 95\% class-stratified origin bootstrap intervals (10,000 resamples); all seed predictions remain paired within each image. They condition on the selected trained states and fixed actions.}\label{tab:official_intervals}
\begin{tabular*}{\textwidth}{@{\extracolsep{\fill}}llrrr@{}}\toprule
Reader & Action & Change (pp) & 95\% interval (pp) & Disagreement (\%)\\
\midrule
HiddenProbe & Perm. & +0.00 & $[+0.00,+0.00]$ & 0.00 \\
HiddenProbe & Phase 1 & -0.23 & $[-0.48,+0.02]$ & 7.28 \\
HiddenProbe & Phase 3 & -3.41 & $[-3.88,-2.95]$ & 22.03 \\
HiddenProbe & Sign & +0.10 & $[-0.20,+0.40]$ & 10.52 \\
ProbeGen & Perm. & +0.00 & $[+0.00,+0.00]$ & 0.00 \\
ProbeGen & Phase 1 & +0.00 & $[+0.00,+0.00]$ & 0.00 \\
ProbeGen & Phase 3 & +0.00 & $[+0.00,+0.00]$ & 0.00 \\
ProbeGen & Sign & +0.00 & $[+0.00,+0.00]$ & 0.00 \\
\bottomrule\end{tabular*}
\end{table*}

\paragraph{Corpus and fixed configurations.}
We reuse the official NFN CIFAR10 SIREN corpus linked by the HiddenProbe
repository \cite{zhou2023nfn,sarangi2026hiddenprobe}. The downloaded archive
contains augmented and original networks; only the 60,000 original
networks are retained. Unique network IDs 0--44,999 train readers,
45,000--49,999 validate them, and 50,000--59,999 form the 10,000-image
test set. Labels match the original CIFAR10 binary data. Each field has
two width-32 sine layers, frequency 30, two coordinate inputs, and three
linear outputs. Field weights remain frozen.

The source commit is \texttt{2eb7070}; full hashes and commands are in the
artifact. Seeds 0, 1, and 2 are fixed for both readers. HiddenProbe uses
128 generated probes, a width-120 SetTransformer with two encoders and
eight heads, dropout 0.1, EMA decay 0.999, batch size 32, head/probe
learning rates $2\times10^{-4}$/$7\times10^{-4}$, 300-step warmup, and
head decay 0.1. The plateau scheduler has factor 0.7, patience 5, and
minimum rate $10^{-6}$. ProbeGen uses 128 probes, width 318, six mixer
layers, a two-layer width-256 per-probe MLP with four outputs, batch
size 64, learning rate $7\times10^{-4}$, zero decay, and plateau factor
0.2 with patience 3 and minimum rate $10^{-6}$. Both preserve the released
linear probe generator (latent width 32, generator width 16) and 60 epochs.
Only clean validation accuracy selects checkpoints and controls scheduling;
the released trainers also log test scores every 500 steps. Intermediate
test scores do not tune the protocol, seed set, or actions.

These are released-configuration audits, not an exact replication claim
for the paper's stated 30-epoch recipe. The released implementations
contain 447,908 HiddenProbe and 579,362 ProbeGen parameters, although the
paper's table lists approximately 0.45M for both. No capacity-matched
comparison is claimed. Native three-seed test accuracy is
$57.82\pm0.40$\% and $57.28\pm0.62$\% (standard deviations), respectively;
the paper reports five-seed means of 58.0\% and 57.3\% with standard
errors 0.2 and 0.7 points. Each selected checkpoint's native validation
and test accuracy is reproduced exactly after loading.

\paragraph{Precision and operation checks.}
HiddenProbe retains upstream BF16 training and native clean inference;
ProbeGen retains FP32. We preserve each loader's output convention:
HiddenProbe adds 0.5 to the raw field output, while ProbeGen adds zero.
The audit holds the generated FP32 probe coordinates fixed, evaluates
targets/actions in FP64, and converts captured responses to FP32 at the
readout boundary. Native BF16 and FP32 clean HiddenProbe predictions
disagree on 2.19--2.37\% of test origins, although their net accuracies
are close. FP32 and FP64-target clean predictions agree for every tested
state. We report each pipeline separately, using the audit clean baseline
for all transformed comparisons.

Per-network fused linear operations match the upstream implementation;
a separately tested batched FP64 renderer checks the fields. An initial
batched FP32 feature implementation failed the native-logit tolerance on
the full test population because bias-addition rounding was amplified
by the readout. Using the original fused operations gives zero native
FP32 logit discrepancy for all six final evaluations. Model weights,
actions, and tolerances are unchanged. The artifact retains the failed
implementation check and the correction record.

\paragraph{Actions and preservation gates.}
Actions use the earlier orbit helper's deterministic hash of transform
seed 2026111203, origin ID, and condition. Each hidden neuron receives
an independent signed one- or three-period bias shift; sign flips and
permutations compensate the following layer. There is one fixed draw
per origin/condition, shared across all reader seeds and methods. Fields
are promoted to FP64 before constructing phase shifts. Each action is
checked on the complete $32\times32$ grid, 128 fixed off-grid points,
and all 128 learned probe coordinates, including direct native-operation
checks at those probes. Every FP64 comparison passes the $10^{-10}$ raw
output gate; the largest observed error is $1.52\times10^{-14}$.
The diagnostic FP32 grid maximum is $8.38\times10^{-6}$.

BF16 is unsuitable as an unqualified function-preservation check here:
a 32-training-field preflight found phase-shift output errors up to
0.2578 raw units (0.1289 in RGB units), despite FP64 preservation near
machine precision. Hence native BF16 scores are not used as the clean
baseline for the transformed FP64 audit. These finite-coordinate checks
are numerical diagnostics, not uniform certificates.

\paragraph{Paired results and retained evidence.}
\Cref{tab:official_seeds,tab:official_intervals} report every seed and
condition. The Phase-1 accuracy interval includes zero, whereas 7.28\%
of HiddenProbe predictions change. Phase-3 accuracy decreases by 3.41
points, and sign changes alter 10.52\% of predictions with a near-zero
net accuracy change. Permutations and output-only ProbeGen have no
observed prediction changes. These results qualify the magnitude of
the adapted-study failure while extending its coordinate-sensitivity
diagnosis to the official corpus.

Intervals use 10,000 class-stratified paired origin resamples, retaining
all three reader predictions and actions within each sampled image.
They are descriptive and conditional on the selected states. Per-seed
variation is reported separately. The artifact retains all six selected
checkpoints, original training logs/configurations, per-origin labels
and logits, numerical errors, common field-parameter digests, and hashes.
All runs used the local RTX 2080 Ti; target fields were not refitted.

\fi
\end{document}